%% file: main.tex
\documentclass{article}

\usepackage{soul}
\usepackage[final]{neurips2022}

\usepackage{nicefrac}
\usepackage[hidelinks]{hyperref}
\usepackage{wrapfig}
\usepackage{microtype}
\usepackage{graphicx}
\usepackage{comment}

\usepackage[FIGTOPCAP]{subfigure} %
\usepackage{soul}
\usepackage[dvipsnames]{xcolor}
\usepackage{booktabs} %
\usepackage{multirow}
\usepackage{soul}
\usepackage{array}
\usepackage[para,online,flushleft]{threeparttable}
\usepackage{floatrow}
\newfloatcommand{capbtabbox}{table}[][\FBwidth]
\usepackage{blindtext}
\usepackage[font=small]{caption} %

\usepackage{natbib}

\hypersetup{
  citecolor=Blue,
  colorlinks=true,
  linkcolor=Blue,
  filecolor=Blue,      
  urlcolor=Blue}

\usepackage{amsmath}
\usepackage{amsfonts}
\usepackage{amssymb}
\usepackage{mathtools}
\usepackage{amsthm}
\usepackage{booktabs}
\usepackage[capitalize,noabbrev]{cleveref}
\usepackage{nccmath}

\usepackage[T1]{fontenc}
\usepackage[utf8]{inputenc}

\usepackage{algorithm2e}
\usepackage{algpseudocode}

\definecolor{b2}{RGB}{51,153,255}
\definecolor{green}{RGB}{80,180,0}
\definecolor{yl}{RGB}{255,80,0}
\definecolor{my_orange}{RGB}{252, 229, 205}
\definecolor{my_green}{RGB}{182, 215, 168}
\definecolor{my_red}{RGB}{244, 204, 203}
\definecolor{my_yellow}{RGB}{255, 229, 153}

\newcommand{\FILM}{{FILM}}

\newcommand\ti[1]{\textit{#1}}

\newcommand\tf[1]{\textbf{#1}}

\newcommand\mf[1]{\mathbf{#1}}

\usepackage[shortlabels]{enumitem}

\newcommand{\probP}{\text{I\kern-0.15em P}}

\theoremstyle{plain}

\theoremstyle{definition}

\theoremstyle{remark}

\renewcommand{\cite}[1]{\citep{#1}}

\everypar{\looseness=-1}

\title{Recovering Private Text in Federated Learning of \\ Language Models}

\author{%
  Samyak Gupta\thanks{The first two authors contributed equally.} \\
  Princeton University\\
  \texttt{samyakg@cs.princeton.edu} \\
  \And
  Yangsibo Huang$^*$ \\
  Princeton University\\
  \texttt{yangsibo@princeton.edu} \\
  \And
  Zexuan Zhong \\
  Princeton University\\
  \texttt{zzhong@cs.princeton.edu} \\
  \And
  Tianyu Gao \\
  Princeton University\\
  \texttt{tianyug@cs.princeton.edu} \\
  \And
  Kai Li \\
  Princeton University\\
  \texttt{li@cs.princeton.edu} \\
  \And
  Danqi Chen \\
  Princeton University\\
  \texttt{danqic@cs.princeton.edu} \\
}

\begin{document}

\maketitle

\vspace{-4mm}
\begin{abstract}
  \input{abstract}

\end{abstract}

\linepenalty=1000 %
\everypar{\looseness=-1} %

\setlength{\lineskiplimit}{0pt}
\setlength{\lineskip}{0pt}
\setlength{\abovedisplayskip}{-0.5ex}
\setlength{\belowdisplayskip}{-0.5ex}
\setlength{\abovedisplayshortskip}{0pt}
\setlength{\belowdisplayshortskip}{0pt}    %

\input{intro}

\input{related}

\input{preliminary}
\input{method_no_freq}

\input{defense}

\input{exp}

\input{analysis}

\input{conclusion}

\newpage
\bibliography{ref}
\bibliographystyle{icml2022}
\setcitestyle{authoryear}

\newpage
\input{checklist}

\newpage
\appendix
\input{appendix}

\end{document}

%% file: abstract.tex
\vspace{-3mm}
Federated learning allows distributed users to collaboratively train a model while keeping each user’s data private. Recently, a growing body of work has demonstrated that an eavesdropping attacker can effectively recover image data from gradients transmitted during federated learning. However, little progress has been made in recovering text data. In this paper, we present a novel attack method {\FILM} for federated learning of language models (LMs). For the first time, we show the feasibility of recovering text from \ti{large batch sizes} of up to 128 sentences. Unlike image-recovery methods that are optimized to match gradients, we take a distinct approach that first identifies a set of words from gradients and then directly reconstructs sentences based on beam search and a prior-based reordering strategy. 
We conduct the {\FILM} attack on several large-scale datasets and show that it can successfully reconstruct single sentences with high fidelity for large batch sizes and even multiple sentences if applied iteratively.
We evaluate three defense methods: gradient pruning, DPSGD, and a simple approach to freeze word embeddings that we propose.  We show that both gradient pruning and DPSGD lead to a significant drop in utility. However, if we fine-tune a public pre-trained LM on private text without updating word embeddings, it can effectively defend the attack with minimal data utility loss. Together, we hope that our results can encourage the community to rethink the privacy concerns of LM training and its standard practices in the future.\protect\footnote{Our code is publicly available at \href{https://github.com/Princeton-SysML/FILM}{https://github.com/Princeton-SysML/FILM}.}

%% file: intro.tex
\vspace{-4.5mm}
\section{Introduction}
\label{sec:intro}
\vspace{-3mm}

\begin{minipage}{0.42\textwidth}
    Federated learning~\cite{mcmahan2016communication} is a method to allow multiple participants to collaboratively train a global model without exchanging their private data. At each step of training, a central server transmits model parameters to every participating client. Each client then computes a model update (i.e., gradients) using its local data and sends it to the server. Finally, the server aggregates all updates---typically by averaging them---and updates the model. Federated learning is actively being considered for privacy-sensitive applications such as virtual mobile keyboards and analysis of electronic health records in hospitals~\cite{li2020federated}.
\end{minipage} \hfill
\begin{minipage}{0.53\textwidth}
    \vspace{-3mm}
    \begin{figure}[H]
        \vspace{-7mm}
        \includegraphics[width=1.02\linewidth]{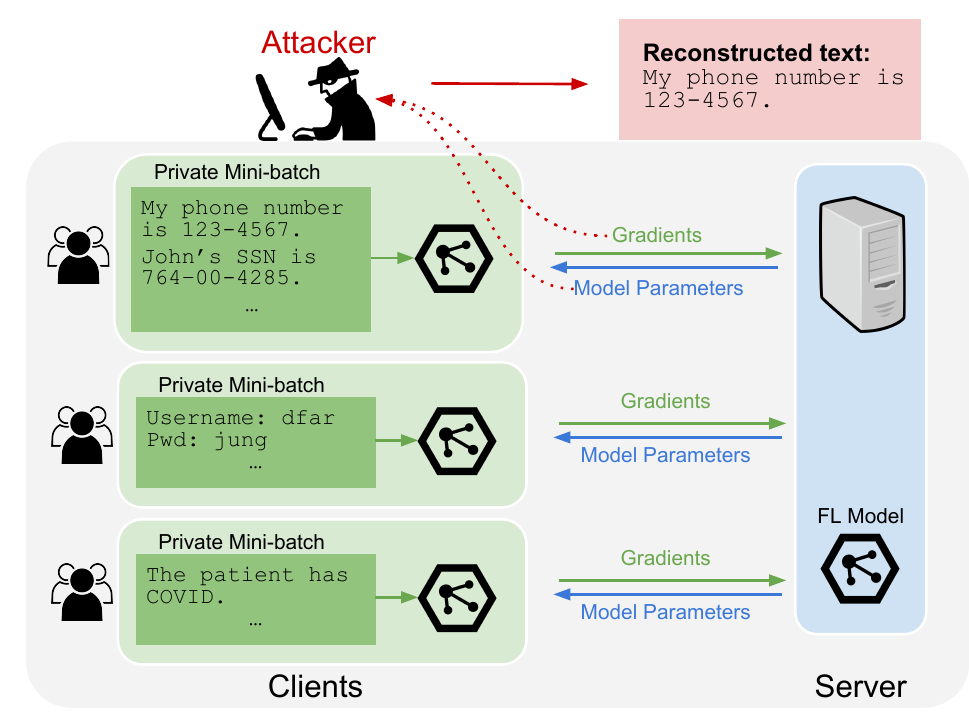}
        \vspace{-6mm}
        \caption{{\small {\FILM} allows an \textit{honest-but-curious} eavesdropper on the communication between the client and the server in federated learning to recover private text of the client.}}
        \label{fig:intro_fig}
    \end{figure}
\end{minipage}

However, recent studies~\cite{zhu2020deep, geiping2020inverting,pustozerova2020information,lyu2020privacy,yin2021see,zhu2021r, huang2021evaluating,gupta2021membership} show that an \textit{honest-but-curious} eavesdropper in federated learning can recover private data of the client (Fig.~\ref{fig:intro_fig}). However, success in recovering text data remains limited to unrealistically small batch sizes (see Sec.~\ref{sec:related-work} for more details).

In this work, we study federated learning of neural language models (LMs) and present a novel attack called FILM (\tf{F}ederated \tf{I}nversion Attack for \tf{L}anguage \tf{M}odels), which can recover private text data from information transmitted during training. For the first time, we demonstrate that an attacker can successfully recover sentences from \ti{large} training batches of up to 128 sentences, making it practical and alarming in real-world scenarios. We focus on language models for two reasons: (1) Transformer-based~\cite{vaswani2017attention} language models have become the backbone of modern NLP systems~\cite{devlin-etal-2019-bert,radford2018improving,radford2019language,brown2020language} and are quickly adapted to a number of privacy-sensitive domains~\cite{alsentzer2019publicly,huang2019clinicalbert,li2020behrt,liu2020federated,kraljevic2021medgpt}; (2) We identify that training LMs is at a higher risk of information leakage by nature because the attacker can possibly leverage the memorization capability of LMs during federated learning.

Our {\FILM} attack deviates from previous gradient-matching approaches for image data that directly optimize in high-dimensional, continuous space, which we find to be ineffective for discrete text inputs and highly sensitive to initializations (Sec.~\ref{sec:exp}).  Instead, we first recover a set of words %
from the word embeddings' gradients~\cite{melis2019exploiting}, and this step provides a set of words that appear in any sentences of the training batch without the frequency information\footnote{We may still estimate the frequency of words based on the magnitude of gradients~\cite{wainakh2021user, fowl2022decepticons}. 
However, later we show that our attack can actually compensate for the lack of frequency information (Sec.~\ref{sec:method_beamsearch}).}. %
We then develop a simple and effective strategy using beam search~\cite{reddy1977speech,russell2002artificial}, which attempts to reconstruct one sentence from the set of  words. This step is crucial as it takes advantage of prior knowledge encoded in pre-trained language models and memorization of training data during federated training. We also further design a token-reordering approach that leverages both language prior and gradient information to refine the recovered sentence. Finally, we show that we can recover multiple sentences from the same training batch by iteratively applying the same procedure.

We evaluate {\FILM} on two language modeling datasets: WikiText-103~\cite{merity2016pointer} and Enron Email~\cite{klimt2004enron}, based on a GPT-2 model~\cite{radford2019language}, using either pre-trained or randomly-initialized weights. We analyze the attack performance with different batch sizes, the number of training data points, and the number of training epochs. Our experiments demonstrate high-fidelity %
recovery of a single sentence feasible, and recovery of significant portions of sentences for training batches of up to $128$ sentences. Furthermore, we find that recovery of multiple sentences from the same batch ($1/3$ for a batch size of 16) is possible through repeated applications of FILM. 

For defense, we first evaluate two previously proposed defense methods, gradient pruning~\cite{zhu2020deep} and Differentially Private Stochastic Gradient Descent (DPSGD)~\cite{abadi2016deep} against the proposed attack and find that both of them suffer substantial utility loss. We then propose a simple method to \ti{freeze} the word embeddings of the model during training to prevent the critical first stage of our attack and find that (1) if we train an LM from the scratch on the private text, it will cause more utility loss; (2) however, if we start from a public LM (e.g., pre-trained GPT-2), it can effectively defend the {\FILM} attack with minimal drop in utility. Together, our work presents a strong attack on federated learning of LM training, and raises privacy concerns in current practices. We hope our new (and simple) proposal of \ti{freezing} word embeddings from a public LM encourages reconsideration of standard practices for training LMs on privacy-sensitive domains.

%% file: related.tex
\vspace{-3mm}
\section{Related work}
\label{sec:related-work}
\vspace{-3mm}

\paragraph{Data recovery from gradients.} Gradients in federated learning have been shown to leak information of private data.   Prior works reconstruct private training images from gradients by treating the reconstruction as an optimization problem~\cite{zhu2020deep, zhao2020idlg, geiping2020inverting, enthoven2021fidel, yin2021see, zhu2021r, jin2021catastrophic, jeon2021gradient}. They iteratively manipulate reconstructed images to yield similar gradients as observed gradients computed on the private images, using regularization terms based on image priors such as total variation~\cite{zhu2020deep, geiping2020inverting} and batch normalization statistics~\cite{yin2021see}.

\vspace{-1mm}
The first attempt to recover text from gradients is \citet{zhu2020deep}, which briefly presented leakage results in masked language modeling~\cite{devlin-etal-2019-bert}. This attack matches target gradients with continuous representations and maps them back to words that are closest in the embedding matrix. \citet{deng2021tag} improve the method by adding a regularization term that prioritizes gradient matching in layers closer to the input data. Both approaches only work with a batch size of 1 (Sec.~\ref{sec:exp}).  Recently, \citet{dimitrov2022lamp} propose a method to minimize the loss of gradient matching and the probability of prior text computed by an auxiliary language model and show the recovery of a batch of 4 sentences with binary classification tasks with a BERT model.  
In addition, ~\citet{boenisch2021curious, fowl2022decepticons} recover text data in federated learning under a stronger threat model: the server is malicious and can manipulate the training model's weights to enable easier text reconstruction.

Our work builds on the insight from \citet{melis2019exploiting} that show the information leakage from the embedding layer by observing non-zero entries in the gradients of the word embedding matrix. Different from their attack which only infers a person or location from a shallow GRU model~\cite{cho2014properties}, we propose a much stronger attack that aims at recovering full sentences from larger language models. We include a high-level comparison of key differences between our approach and prior approaches in Table~\ref{tab:comparison_prior} in the appendix.

\vspace{-3mm}
\paragraph{Memorization in language models.}
Our method is also inspired by the work related to memorization of training data in language models. Recent studies~\cite{carlini2019secret, thakkar2020understanding, song2020information, zanella2020analyzing, carlini2021extracting, bender_dangers_2021} notice that large language models can memorize their training dataset and thus can be prompted to output specific sensitive information. 
While these previously demonstrated memorization attacks try to recover {\it a small subset of} sensitive text from a large training corpora, our attack aims at recovering sentences from {\it every} mini-batch in federated learning. %
We stress that our attack is significantly more powerful than the previous works studying memorization.

\vspace{-0.1in}
\paragraph{{Defenses against gradient inversion attacks for text data.}} Cryptographic approaches such as encrypting gradients~\cite{bonawitz2016practical} or encrypting the data and model~\cite{phong18} can guarantee secure training in a federated learning setting against gradient inversion attacks. 
However, practical deployment of these approaches often slows down model training and requires special setup. A few efficient methods are proposed to defend gradient inversion attacks are for text data.
\citet{zhu2020deep} mention briefly adding differentially private noise to the gradients~\cite{abadi2016deep}, or setting gradients of small magnitudes to zero (i.e., gradient pruning), but these defenses typically hurt the accuracy
of the trained models~\cite{li2021large, yu2021differentially}. \citet{huang2020instahide} propose InstaHide that
encodes each input training image to the neural network using a random pixel-wise mask and the MixUp data augmentation ~\cite{zhang2017mixup}. The idea is later extended to language understanding tasks~\cite{huang2020texthide}: instead of `hiding' the input word embeddings, they perform MixUp on \texttt{[CLS]} tokens of BERT models.

%% file: preliminary.tex
\vspace{-2mm}
\section{Preliminaries}
\vspace{-3mm}
In this section, we describe relevant background of language models, federated learning, as well as the threat model of our proposed attack method.

\vspace{-3mm}
\subsection{Language Modeling}
\label{sec:prelim_lm}
\vspace{-2mm}
Language modeling is a core task in natural language processing and has been used as a building block of state-of-the-art pipelines~\cite{radford2018improving,brown2020language}.
Given a sequence of $n$ tokens $\mathbf{x}=\{x_1,x_2,...,x_n\}$,
the language modeling task is to estimate the probability distribution of $\probP(\mathbf{x})$:
\begin{equation}
    \small
    \log \probP_{\theta}(\mathbf{x}) = \sum_{i=1}^{n} \log \probP_{\theta}(x_i|x_1,...,x_{i-1}).
    \label{eq:likelihood}
\end{equation}
Modern neural language models are parameterized by RNNs~\cite{mikolov2010recurrent} or Transformers~\cite{vaswani2017attention}, which consist of millions or billions of parameters (denoted $\theta$). A language model first maps $\mathbf{x}$ to a sequence of vectors via a word embedding matrix $\mathbf{W}\in \mathbb{R}^{|\mathcal{V}| \times d}$, where $\mathcal{V}$ is the vocabulary and $d$ is the hidden dimension. After computing the hidden representation $\mf{h}_i$ conditioned on $x_1, \ldots, x_{i-1}$, the model predicts the probability of the next token as:
\begin{equation}
    \small
    \probP_{\theta}(x_i|x_1,...,x_{i-1}) = \frac{\exp(\mathbf{h}_i^\top \mathbf{W}_{x_i})}{\sum_{j \in \mathcal{V}} \exp(\mathbf{h}_i^\top\mathbf{W}_j)}.
\end{equation}

\vspace{-2mm}
\subsection{Federated Learning}
\vspace{-2mm}

Federated learning (FL)~\cite{mcmahan2016communication} is a communication protocol for training a shared machine learning model on decentralized data. An FL system usually involves $N$ clients: $c_1, c_2, \cdots, c_N$  and a central server $s$, where the clients wish to collaboratively train a neural network $f_\theta$ with their private data $\mathcal{D}_1, \mathcal{D}_2, \cdots, \mathcal{D}_N$, under the coordination of the server. The training process optimizes $\theta$ (model parameters) using a loss function $\mathcal{L}$ and runs for $T$ iterations. This work focuses on federated learning of language models, where clients can start from a randomly-initialized neural network, or a pre-trained public model~\cite{hard2018federated}. %
At each iteration $t \in [T]$, an individual client $c_i, i \in [N]$ computes $\nabla_{\theta^t}\mathcal{L}_{\theta^t}(\mathcal{B}_i)$, the gradients of the current model parameters ${\theta^t}$ on a randomly-sampled private data batch $\mathcal{B}_i \subset \mathcal{D}_i$, $|\mathcal{B}_i| = b$, and shares the gradients with the server $s$. The server then aggregates gradient updates from all clients and updates the model:
\vspace{-1mm}
\begin{equation}
    \small
    \theta^{t+1}={\theta^{t}-\eta \sum_{i=1}^{N} \nabla_{\theta^t} \mathcal{L}_{\theta^{t}}\left(\mathcal{B}_i\right)},
\label{eq:fl_update}
\vspace{-1mm}
\end{equation}
where $\eta$ is the learning rate.
Finally, the updated parameters $\theta^{t+1}$ are broadcast to individual clients.\footnote{Note that Eq.~\ref{eq:fl_update} uses SGD for an easier demonstration for an update step in federated learning. People usually use Adam~\cite{kingma2014adam} in real-world training of neural language models.}
\vspace{-1mm}
\subsection{Threat Model}
\label{sec:threat_model}
\vspace{-2mm}

\paragraph{Adversary’s capabilities.}  We consider an honest-but-curious adversary who is eavesdropping on the communication between the central server $s$ and an arbitrary client $c_i, i \in [N]$ in the federated training of a language model parameterized by $\theta$, as described above. An adversary in this scenario has white-box access to 1) the gradients $\nabla_{\theta^t}\mathcal{L}_{\theta^t}(\mathcal{B}_i)$ sent by the client $c_i$, and 2) the model parameters $\theta^t$ (Fig.~\ref{fig:intro_fig}), including the vocabulary $\mathcal{V}$ and the embedding matrix $\mathbf{W}$. Note that the adversary can inspect these information at \ti{any} training iterations $t \in [T]$.

\vspace{-2mm}
\paragraph{Adversary’s objective.} The adversary aims to recover {\it at least one} sentence from the private training data batch $\mathcal{B}_i$ based on the information they observe, as recovering a single sentence is already sufficient to break the privacy guarantee of federated learning. Since each training batch consists of randomly selected sentences from the private training dataset, if the adversary could recover at least one sentence from each training batch during the whole training procedure, they would be able to recover a considerable fraction of the original training dataset. Additionally, the attacker may also apply the attack on a single batch multiple times to recover more sentences from the batch (Sec.~\ref{sec:multi-sentences}).
The strength of an attack will be measured by the similarity between the recovered sentence and its corresponding private sentence in the original batch.

%% file: method_no_freq.tex
\vspace{-2mm}
\section{Proposed Attack: {\FILM}}
\label{sec:attack_method}
\vspace{-2mm}

\begin{figure*}[t]
    \vspace{-6mm}
    \centering
    \makebox[\textwidth][c]{
    \includegraphics[width=1.02\textwidth]{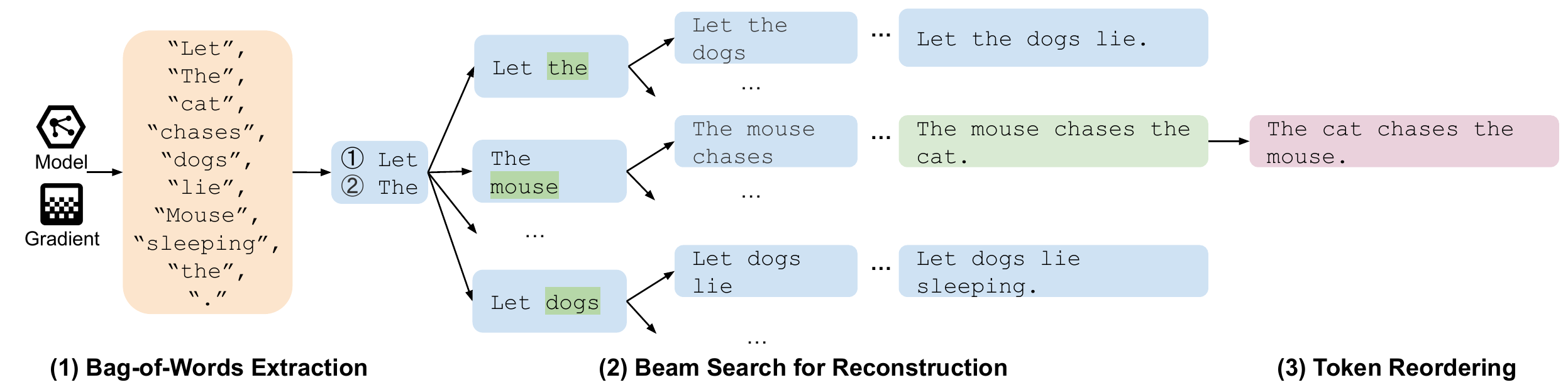}}
    \vspace{-6mm}
    \caption{
        Illustration of our pipeline for a batch size of 2.
        We first recover a bag of words from the observed gradients, and then generate candidate sentences using beam search from the set of words, retaining top beams at every step according to a beam scoring function. Finally, we re-order phrases and words in the best candidate.}
    \label{fig:beamsearch}
\end{figure*}

\subsection{Overview}
\vspace{-2mm}

In this section, we present a novel attack method {\FILM}, which can construct sentences from \ti{any} training batches ($|\mathcal{B}_i| \leq 128$). Our method consists of three steps:
\vspace{-1mm}

(1) Inspired by \citet{melis2019exploiting}, we first recover a set of words that may appear in a private batch ($\mathcal{B}_i$) from gradients (Sec.~\ref{sec:method_word}). While their goal is to identify individual words from gradients, our approach is able to recover the order of words in a sentence.
\vspace{-1mm}

(2) We attempt searching for full sentences using beam search using the set of words, based on the probability distribution $\probP_{\theta^t}(x)$, which leverages either the language prior from a pre-trained language model ($t=0$), or the memorization capability of the LM during training ($t>0$) (Sec.~\ref{sec:method_beamsearch}).\footnote{Recall that the attacker can inspect the gradients at any step $t \in [T]$ during the $T$-epoch training. 
An attacker performing our attack at a larger $t$ may have much stronger results than for a smaller $t$, as we later show in Sec.~\ref{sec:analysis}.%
}
\vspace{-1mm}

(3) Finally, we design a prior-based scoring function and demonstrate that we can further improve the quality of recovered sentences by performing a simple and lightweight reordering step (Sec.~\ref{sec:method_finetune}).

\vspace{-1mm}

We show that the above pipeline is capable of recovering single sentences from large training batches successfully. In Sec.~\ref{sec:multi-sentences}, we discuss that we can iterate the above procedure to recover multiple sentences from the same training batch, which makes our attack even stronger. In the following, we describe our attack method in detail. To better illustrate our attack's functionality, we provide a simplified step-by-step example that recovers a batch of 2 simple sentences in Fig.~\ref{fig:beamsearch}.

\vspace{-3mm}
\subsection{Bag-of-Words Extraction}
\label{sec:method_word}
\vspace{-1mm}
{\small
    \textbf{Input:} $\nabla_{\theta^t}\mathcal{L}_{\theta^t}(\mathcal{B}_i)$,  gradients of the language model on the private batch of sentences.

    \vspace{-1mm}
    \textbf{Output:} bag of words $B$: \texttt{\small \{``Let", ``The", ``cat", ``chases", ``dogs", ``lie", ``mouse", ``sleeping", ``the", ``."\}}. }
Our first step is to extract a bag of words from $\nabla_{\theta^t}\mathcal{L}_{\theta^t}(\mathcal{B}_i)$, the gradients observed by an eavesdropper  at the $t$-th iteration of training.
We follow the method outlined by \citet{melis2019exploiting} and recover a set of words $B = \{w_1, w_2, \dots, w_m\}$  by considering the non-zero rows of the word embedding gradients $\nabla_{\mf{W}^t}\mathcal{L}_{\theta^t}(\mathcal{B}_i)$. %
Additionally, we are also able to recover the maximum length of sentences in the batch by analyzing the non-zero rows of position embeddings gradients $\nabla_{\mf{P}^t}\mathcal{L}_{\theta^t}(\mathcal{B}_i)$.\footnote{We assume that sentences in one batch vary in length. A standard practice is to pad shorter sentences.} Note that this step only determines the set of words but not their \ti{frequency}. {In practice, we find that augmenting our beam search with an n-gram penalty (Sec.~\ref{sec:method_beamsearch}) can  compensate for this lack of information.} %

\vspace{-3mm}
\subsection{Beam Search for Sentence Reconstruction}
\label{sec:method_beamsearch}
\vspace{-2mm}

{\small
    \textbf{Input:}  $B$, the bag of private words recovered by Sec.~\ref{sec:method_word}. %
    \vspace{-2mm}
    
    \textbf{Output:} a recovered sentence from the batch, \texttt{\small ``The mouse chases the cat."}.
}
\vspace{-1mm}

The second stage of our attack is to recover \ti{one} sentence based on the extracted bag of words $B$ {and the learned probability distribution of text $\probP_{\theta^t}(\mathbf{x})$} at training step $t$. %
Since it is intractable to generate text directly using $\probP_{\theta^t}(\mathbf{x})$, we use the beam search~\cite{reddy1977speech, russell2002artificial}, which is a greedy auto-regressive algorithm to generate sequences in a step-by-step manner with a trained language model. We provide a detailed algorithm in Appendix~\ref{app:beam_search}.

To initiate the beam search, we first find a set of words that can be chosen as starting tokens of sentences. A straightforward method is to select words that begin with a capital letter from $B$. For example, we would select \texttt{``Let"} and \texttt{``My"} as possible prompts for the illustrated batch. Then at each beam search step, we keep the top $k$ possible beam states, find the $k$ most likely next words to each beam by using the language model (see procedure \textsc{Top($\cdot$)} in Appendix~\ref{app:beam_search}, Algorithm~\ref{alg:beamsearch}), and keeps the top $k$ new beam states. The beam tree continues until the generated sentence reaches a specified length (we use $40$ words), and outputs the text that has the overall highest probability (see Fig.~\ref{fig:beamsearch}). %

As the bag of words $B$ does not contain information on the frequency of each word in the sentence, and beam search is prone to generating repetitive content \cite{Holtzman2020The}, we also include a penalty for repeated $n$-grams~\cite{vijayakumar2016diverse} in a sentence (see line \ref{alg:scoring} in procedure \textsc{Top($\cdot$)}). Noticeably, we compare the performance of our attack with an oracle scenario where the attacker knows the frequency of each word in Fig.~\ref{fig:exp_ablation_attack}, and find that adding an n-gram penalty ($n = 2$) can achieve performance comparable without frequency information.

\vspace{-3mm}
\subsection{Prior-Guided Token Reordering}
\vspace{-2mm}
\label{sec:method_finetune}
{\small
    \textbf{Input:}  a recovered sentence from Sec.~\ref{sec:method_beamsearch}, \texttt{\small ``The mouse chases the cat."}

    \vspace{-2mm}
    \textbf{Output:} a reordered sentence, \texttt{\small ``The cat chases the mouse."}
}
\vspace{-1mm}

By construction, beam search is a greedy method that generates sentences from left to right. Despite that it generally produces fluent sentences by maximizing $\probP_{\theta^t}(\mathbf{x})$, it does not leverage the \ti{gradients} information which is the main signal in previous image attack methods. We also notice that beam search  may recover the general structure of the sentence, while failing to recover the order of specific phrases or words.  Thus, the final stage of our attack is to reorder tokens in the extracted sentence from the previous step, with the objective to improve its quality measured by a prior-based scoring function. The scoring function leverages both the perplexity and the gradient norm.
Formally, given a  sentence  $\mathbf{x}$, its prior score $\mathcal{S}_{\theta}(\mathbf{x})$ is defined as:
\vspace{2mm}
\begin{equation}
    {\small
    \mathcal{S}_{\theta}(\mathbf{x}) = \underbrace{\exp\left\{-\frac{1}{n} \log \probP_\theta(\mathbf{x})\right\}}_{\text{Perplexity}} + \beta
    \underbrace{\|\nabla_{\theta}\mathcal{L}_{\theta}(\mathbf{x})\|}_{\text{Gradient Norm}}},
    \label{eq:reorder_score}
    \vspace{1mm}
\end{equation}
where $\log \probP_\theta(\mathbf{x})$ is the log-likelihood of the sentence (Eq.~\ref{eq:likelihood}) according to the model parameters $\theta$, and $\beta$ is a hyperparameter that controls the importance of gradient norm.  Appendix~\ref{sec:finetune_stats} provides empirical analysis that supports the design of this scoring function.  We conduct two additional steps to reorder both phrases and tokens:
\paragraph{Phrase-wise reordering.} We first preprocess the sentence $\mathbf{x}$ by removing its redundant segments\footnote{We notice that the sentences we recover from the previous step may not naturally finish after the beam search generates a punctuation. To address this problem, we preprocess each recovered sentence $\mathbf{x}$ by removing segments after its first punctuation, as long as the resulting sentence $\mathbf{x'}$ satisfies $\mathcal{S}_\theta(\mathbf{x'}) < \mathcal{S}_\theta(\mathbf{x})$.}. We then adjust the preprocessed sentence by reordering the phrases inside it. At each iteration, we generate multiple candidates by first cutting the sentence at $p$ positions using the strategy in \citet{malkin2021studying}, and then permuting the phrases between the cuts to form a new sentence. We then select the best candidate measured by $\mathcal{S}_{\theta}(\cdot)$. This stage terminates after 200 steps. %

\vspace{-4mm}
\paragraph{Token-wise reordering.} We then reorder individual tokens. At each reordering step, we generate candidates using the following token-wise operations: 1) randomly swaps two tokens in the sentence, or 2) randomly deletes a token from the sentence, or 3) randomly inserts a token from the bag of words (recovered by Sec.~\ref{sec:method_word}) into the sentence. We then select the best candidate measured by $\mathcal{S}(\cdot)$ and move to the next iteration. This stage terminates after 200 steps. %

\vspace{-3mm}
\subsection{Recovering Multiple Sentences}
\label{sec:multi-sentences}
\vspace{-3mm}
The previous steps only consider the recovery of a single sentence from a training batch. In this step, we describe an extension of our attack which is able to support recovery of multiple sentences from the same batch. We first perform beam search as described in Sec.~\ref{sec:method_beamsearch}. We store the results and repeat the beam search again, except this time applying an additional $n$-gram penalty based on the results of the previous search. Over many repetitions of this procedure, we are able to build a set of candidate recoveries for the original batch. Finally, we use the reordering method described in Sec~\ref{sec:method_finetune} to improve each candidate.

%% file: defense.tex
\vspace{-3mm}
\section{Defending Against the {\FILM} Attack}
\label{sec:defenses}

\vspace{-3mm}

We evaluate previously proposed defense methods, gradient pruning~\cite{zhu2020deep} and DPSGD~\cite{abadi2016deep} and find that both are less effective in defending our attack (Sec.~\ref{sec:exp_defense}).

\par To defend against the {\FILM} attack, we propose a defense method which simply \ti{freezes the word embeddings} of the model during training. Our key insight is to focus on preventing the first step of the attack, i.e., the recovery of the bag-of-words from word embedding gradients.  If the attack fails at this step, the later stages of the attack are no better than beam search over the entire vocabulary as with the prior attack~\cite{carlini2021extracting}.  Similarly, previously proposed gradient inversion attacks~\cite{zhu2020deep,fowl2022decepticons} %
for smaller batch sizes also depend on this step. By freezing the word embeddings during training, it is able to completely prevent the recovery of bag-of-words. In other words, once we prevent the transmission of gradients of word embeddings during training, it is no longer possible to recover the bag-of-words.

We consider the settings for both 1) training an LM from the scratch; 2) continuing training from a public, pre-trained LM (e.g., GPT-2) on the private text, when freezing the word embeddings. As we later show in Sec.~\ref{sec:exp_defense}, the first setting causes more utility loss (because the word embeddings are randomly initialized) while the latter setting provides the best privacy-utility trade-off. Since updating word embeddings is a common practice in training LMs, we suggest researchers and practitioners consider training from a public LM and freezing word embeddings in privacy-sensitive applications.

%% file: exp.tex
\renewcommand{\arraystretch}{0.85}
\sethlcolor{my_green}

\vspace{-3mm}
\section{Experiments}
\label{sec:exp}
\vspace{-3mm}

\subsection{Setup}
\label{sec:exp_setup}

\vspace{-2mm}
\paragraph{Model and datasets.} We evaluate the proposed attack with the GPT-2 base (117M parameters) model~\cite{radford2019language}  on two language modeling datasets, including WikiText-103~\cite{merity2016pointer}  and the Enron Email dataset~\cite{klimt2004enron}. Both datasets are publicly available for research uses. We choose WikiText-103 because it is commonly used in language modeling research, and Enron Email because it consists of private email messages that contain abundant private information such as individuals' names, addresses, and even passwords. Note that the GPT-2 model is not trained on any Wikipedia data, and very unlikely on the Enron Email data.

We evaluate our attack on a subset of WikiText-103 and Enron Email. After preprocessing (more details provided in Appendix~\ref{app:exp}), we have 203,456 sentences from the WikiText-103 dataset and 31,797 sentences from the Enron Email dataset. 

\vspace{-3mm}
\paragraph{Training and attack settings.} Following previous studies~\cite{zhu2020deep, geiping2020inverting, yin2021see, huang2021evaluating} for gradient inversion attacks, our evaluation considers a federated learning setting with a single server and a single client.\footnote{This setting is often accepted as an adequate stand-in for federated learning as synchronous federated learning with $N$ clients, each with $b$ samples per batch, is (assuming I.I.D. data) functionally equivalent to training a model with a batch size of $N\times b$.} Unless otherwise noted, we train the model on these sentences for $90,000$ iterations using an initial learning rate of $1 \times 10^{-5}$, with a linearly decayed learning rate scheduler. All models were trained using early stopping, i.e., models were trained until the loss of the model on the evaluation set increased.   %
We note that the running time of our algorithm is quite fast, and we can recover a single sentence in under a minute using an Nvidia 2080TI GPU. %

\vspace{-3mm}
\paragraph{Evaluation metrics.} We use the following metrics to evaluate the attack performance: (a) 
\tf{ROUGE}~\cite{lin-2004-rouge} is a set of metrics for evaluating summarization of texts as well as machine translation. We use ROUGE-1/2/L F-Scores to evaluate the similarity between the recovered and the original sentences, following \cite{deng2021tag, dimitrov2022lamp}. More specifically, ROUGE-1/2 refer to the overlap of unigrams and bigrams between the recovered and the original text respectively. ROUGE-L measures the longest matching subsequence. For ablation studies we only show ROUGE-L as it is more representative of significant leakage than ROUGE-1 or ROUGE-2. (b) We also propose to use \tf{named entity recovery ratio (NERR)} as the percentage of original named entities that can be perfectly recovered. Since named entities usually contain sensitive information (e.g., names, addresses, dates, or events), NERR measures how well the attacker recovers such information, ranging from a complete mismatch (NERR = 0) to a perfect recovery (NERR = 1). %

\begin{table}[t]
  \vspace{-12mm}
  \centering
  \setlength{\tabcolsep}{2pt}
  \tiny
  \begin{threeparttable}
    \begin{tabular}{p{1.8cm}|p{5.6cm}p{6.2cm}}
      \toprule
      {\bf \scriptsize Attack \& }      & \multirow{2}{*}{\bf \scriptsize Original Sentence}                                                                                                                                                  & \multirow{2}{*}{\bf \scriptsize Best Recovered Sentence}         
      \\
      {\bf \scriptsize Batch Size $b$ } &                              
      \\
      \midrule
      \multicolumn{3}{c}{\bf \scriptsize WikiText-103}                                                                               \\
      \midrule
      \citet{zhu2020deep}, $b = 1$ & \texttt{As Elizabeth Hoiem explains, "The most English of all Englishmen, then, is both king and slave, in many ways indistinguishable from Stephen Black.}                                         & \texttt{thelesshovahrued theAsWords reporting the Youngerselagebalance, mathemat mathemat mathemat reper arrangpmwikiIndia Bowen perspectoulos subur, maximal}                                                         
      \\
      \midrule
     \citet{deng2021tag}, $b = 1$     & \texttt{Both teams recorded seven penalties, but Michigan recorded more penalty yards.} & 
                                        \texttt{\sethlcolor{my_green}\hl{Both} \sethlcolor{my_green}\hl{recorded} \sethlcolor{my_green}\hl{teams} \sethlcolor{my_green}\hl{seven} \sethlcolor{my_green}\hl{but Michigan} to recorded penalties 40 \sethlcolor{my_green}\hl{more penalty} outstanding	}
      \\
      \midrule
      \FILM, $b = 1$                    & \texttt{The short@-@tail stingray forages for food both during the day and at night.}                                                                                                               & \texttt{\sethlcolor{my_green}\hl{The short@-@tail stingray forages for food both during the day and at night.}}                                                       
      \\
      \midrule

      \FILM, $b = 16$                   & \texttt{A tropical wave organized into a distinct area of disturbed weather just south of the Mexican port of Manzanillo, Colima, on August 22 and gradually moved to the northwest.}               & \texttt{Early \sethlcolor{my_green}\hl{on} September \sethlcolor{my_green}\hl{22}, an \sethlcolor{my_green}\hl{area of disturbed weather} \sethlcolor{my_green}\hl{organized into} \sethlcolor{my_green}\hl{a tropical wave}, which \sethlcolor{my_green}\hl{moved to the northwest} \sethlcolor{my_green}\hl{of} \sethlcolor{my_green}\hl{the} \sethlcolor{my_green}\hl{area}, and then \sethlcolor{my_green}\hl{moved} \sethlcolor{my_green}\hl{into} \sethlcolor{my_green}\hl{the} north and south@-@to \sethlcolor{my_green}\hl{the} northeast.} %
      \\
      \midrule

      \FILM, $b = 128$                  & \texttt{A remastered version of the game will be released on PlayStation 4, Xbox One and PC alongside Call of Duty: Infinite Warfare on November 4, 2016.}                                          & \texttt{At the time of writing, \sethlcolor{my_green}\hl{the game} has been \sethlcolor{my_green}\hl{released on PlayStation 4, Xbox One}, \sethlcolor{my_green}\hl{PlayStation 3}, \sethlcolor{my_green}\hl{and PC}, with the \sethlcolor{my_green}\hl{PC} version being \sethlcolor{my_green}\hl{released} in North America \sethlcolor{my_green}\hl{on November} 18th, 2014.}                                                                                                                                                                     %
      \\
      \midrule
      \multicolumn{3}{c}{\bf \scriptsize Enron Email}                                                                        \\
      \midrule
      \citet{zhu2020deep}, $b = 1$     & \texttt{Don and rogers have decided for cost management purposes to leave it consolidated at this point.}                                                                                           & \texttt{dj "... Free, expShopcriptynt)beccagressive Highlands andinos Andrea Rebell impacts}                                                                                                                      \\
      \midrule
      \citet{deng2021tag}, $b = 1$     & \texttt{We should not transfer any funds from tenaska iv to ena.} & 
                                        \texttt{\sethlcolor{my_green}\hl{We should} \sethlcolor{my_green}\hl{transfer any funds} not \sethlcolor{my_green}\hl{from tenaska iv to} iv en happens	}
      \\
      
      \midrule
      
      \FILM, $b = 1$                    & \texttt{Volume mgmt is trying to clear up these issues.}                                                                                                                                            & \texttt{\sethlcolor{my_green}\hl{Volume mgmt is trying to clear up these issues.}}                                                                                                                                                                                                                                                                                                                                                                                                                                                                   %
      \\

      \midrule
      \FILM, $b = 16$                   & \texttt{Yesterday, enron ousted chief financial officer andrew fastow amid a securities and exchange commission inquiry into partnerships he ran that cost the largest energy trader \$35 million.} & \texttt{\sethlcolor{my_green}\hl{Yesterday, enron ousted} its \sethlcolor{my_green}\hl{chief financial officer}, \sethlcolor{my_green}\hl{andrew fastow}, \sethlcolor{my_green}\hl{amid a securities and exchange commission inquiry into partnerships he ran that cost the} company \sethlcolor{my_green}\hl{\$35 million} in stock and other financial assets.}                                                                                                                                                                                    %
      \\
      \midrule

      \FILM, $b = 128$                  & \texttt{Yesterday, enron ousted chief financial officer andrew fastow amid a securities and exchange commission inquiry into partnerships he ran that cost the largest energy trader \$35 million.} & \texttt{\sethlcolor{my_green}\hl{Yesterday, enron ousted chief financial officer andrew fastow amid a securities and exchange commission inquiry into partnerships he ran that} he said \sethlcolor{my_green}\hl{cost the} company more than \$1 billion in stock and other assets.}                                                                                                                                                                                                                                                                 %
      \\
      \bottomrule
    \end{tabular}
    \vspace{-2mm}
    \caption{Performance of~\citet{zhu2020deep}'s and ~\citet{deng2021tag}'s with batch size = 1 (more in Appendix~\ref{app:zhu-results}) and \FILM~(ours) with different batch sizes. We show the best recovered sentence among 20 tested batches for each batch size (see Fig.~\ref{fig:eval_bs} for average results).
      Text in {\hl{green}} represents successfully recovered phrases and words.}
    \label{tab:attack_example}
  \end{threeparttable}
\end{table}

\vspace{-2mm}
\subsection{Performance of Our Attack}
\label{sec:base_res}

\begin{figure}[t]
  \centering
  \vspace{-16mm}
  \includegraphics[width=\linewidth]{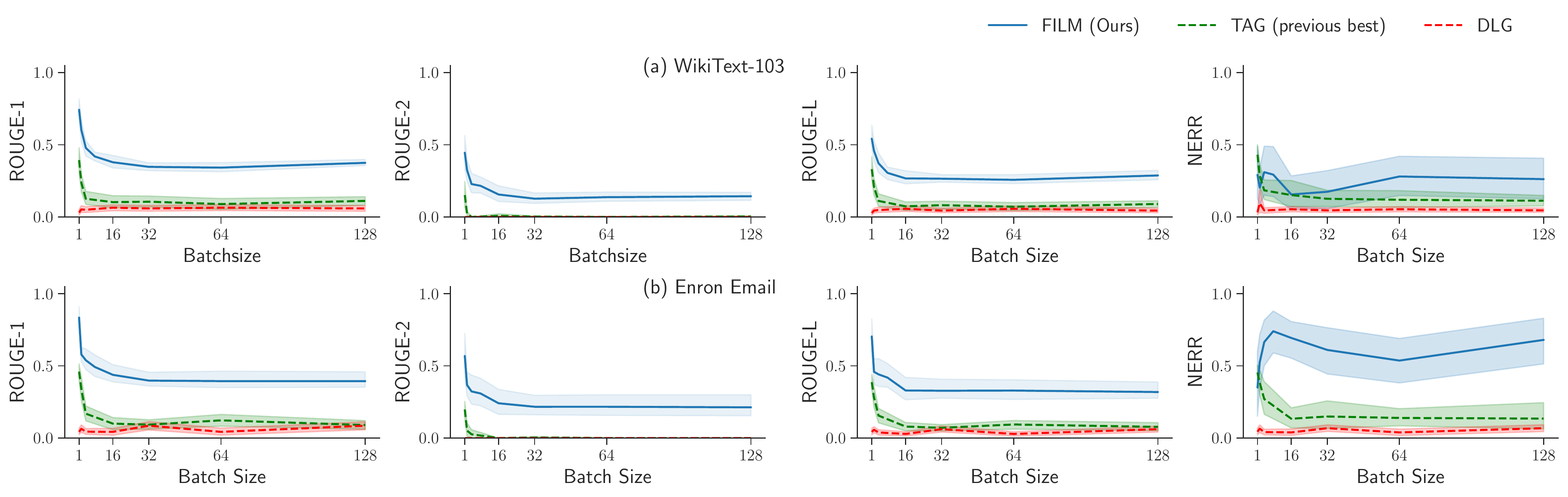}
  \vspace{-8mm}
  \caption{Recovery performance for various batch sizes on WikiText-103 (a) and Enron Email (b). Solid lines indicate the average F-Scores of recoveries out of 20 tested mini-batches. The attack is overall weaker for larger batch sizes (see Table~\ref{tab:main_table} for quantitative results). Moreover, the attack is stronger for the Enron Email dataset.
  }
  \label{fig:eval_bs}
\end{figure}

\vspace{-2mm}
\paragraph{Scaling to large training batches.} We first compare the performance with respect to different batch sizes. In this setting, we start from a pre-trained GPT-2 model and fine-tune it on both datasets respectively (WikiText-103 and Enron Email). 
Fig.~\ref{fig:eval_bs} shows the performance of our attack using  different batch sizes with the WikiText-103 dataset and the Enron Email dataset. We observe that for a batch size of 1, all four scores (ROUGE-1, ROUGE-2, ROUGE-L, NERR) are close to 1, indicating a near-perfect recovery. We observe that as the batch size increases, the quality of the recovery by ROUGE-1, ROUGE-2, and NERR start to decrease. We attribute this drop in performance to the bigger set of words in the same training batch, as well as the growing search space of possible word orderings. We also observe that the Enron Email dataset is more susceptible to attack.

Table~\ref{tab:attack_example} shows that the approach of ~\citet{zhu2020deep} fails to recover the original sentence for batch size $b=1$. We find that the gradient matching method in \citet{zhu2020deep} is  sensitive to initialization and it can perform  better recovery when using a good initialization (see Appendix~\ref{app:zhu-results}). ~\citet{deng2021tag} improves over ~\cite{zhu2020deep} by prioritizing the matching of  gradients of layers that are closer to the input. However, Fig.~\ref{fig:eval_bs} shows the NERR and ROUGE scores sharply degrade for larger batch sizes. On the contrary, our approach is able to recover a large portion of the original sentences on both datasets even when the batch size is 128. This demonstrates the superior effectiveness of leveraging the knowledge encoded in pre-trained language models and its memorization ability to recover sentences, compared to directly performing optimization in a high-dimensional continuous space.

\vspace{-3mm}
\paragraph{Attack performance at different training iterations.} 
We evaluate the attack performance of our approach on the models trained at different numbers of training iterations $t$, using the WikiText-103 dataset. Specifically, we vary the number of training iterations $t \in \{0,10000, \dots, 90000\}$ and fix the batch size for training $b = 16$ and dataset size $N=200,000$. The batch size for attack is $1$.

Previous gradient matching attacks on image data suggests that well-trained models are harder to attack~\cite{geiping2020inverting} due to the shrinkage of gradient norm as training progresses.
However, as shown in Fig.~\ref{fig:attack_t} (see qualitative examples in Table~\ref{tab:attack_iteration}), we find that well-trained models are more vulnerable to our approach because the model is able to memorize data during training.

\vspace{-2mm}
\paragraph{Pre-trained vs randomly-initialized models.}
We additionally compare the attack performance for training that starts from a pre-trained model (GPT-2 in our case) and a randomly-initialized model (we use the same GPT-2 architecture but re-initialize all the weights randomly). 
We find that starting from a pre-trained model is more susceptible to attack than training from scratch as shown in Fig.~\ref{fig:attack_t}. However, the gap in performance gradually decreases as the number of iterations increases. 

\vspace{-1mm}
\begin{minipage}{0.59\textwidth}
As shown in Fig.~\ref{fig:attack_t}, a pre-trained model which has not been fine-tuned on client data can recover sentences with an average ROUGE-L score of 0.41, indicating that prior knowledge encoded in pre-trained models may help recover sentences. When the model is continued to train on Wikitext-103, the average score increases to 0.53, an almost 30\% increase. Our results suggest that leveraging the memorization in the model yields a more powerful attack than only using the language prior encoded in the pre-trained model. We note that we would expect stronger attack performance (especially for smaller training epochs) when attacking larger-capacity models, as stronger models can memorize training instances easily. 
We report more results for $t=0$ with different batch sizes in Table~\ref{tab:attack_t_0} in  Appendix~\ref{app:t0-results}.
\end{minipage}
\hfill
\begin{minipage}{0.39\textwidth}
\begin{figure}[H]
\vspace{-3mm}
    \centering
    \includegraphics[width=0.9\linewidth]{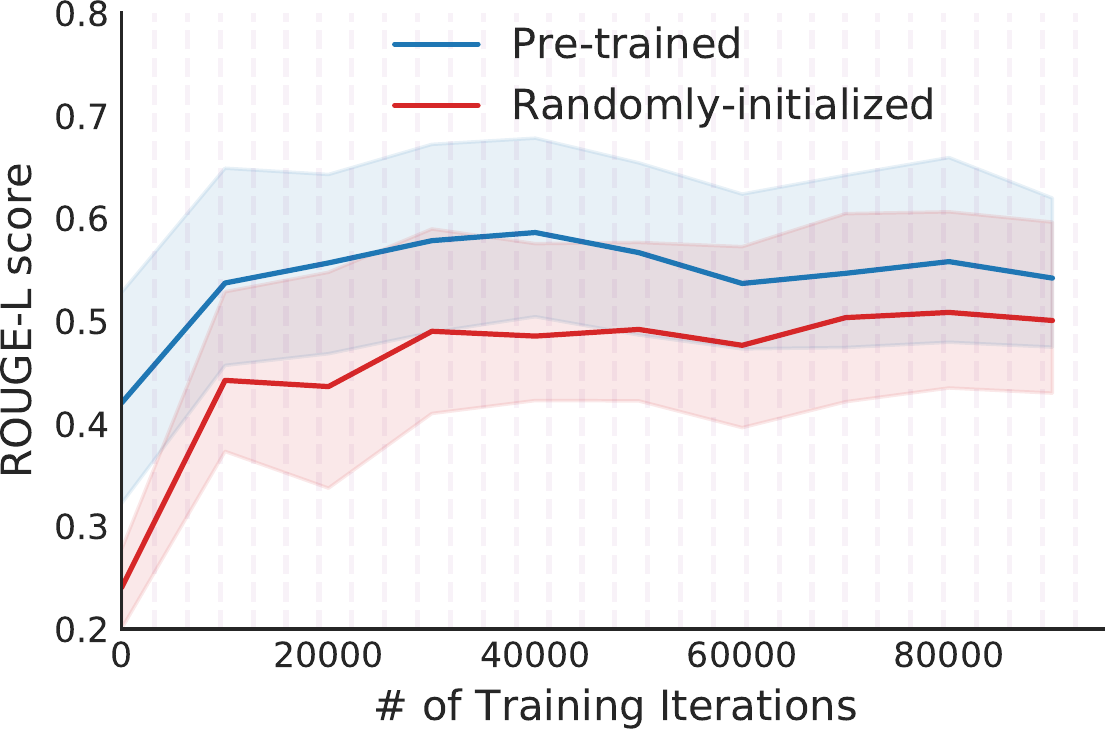}
    \vspace{-3mm}
    \caption{Attack performance over training iterations $t$ for pre-trained  or  randomly-initialized model on WikiText-103 (batch size for the attack is $1$). Dashed lines indicate when an epoch was completed. See  Appendix~\ref{app:t0-results} for qualitative results.}
    \label{fig:attack_t}
\end{figure}
\end{minipage}

\vspace{-3mm}
\paragraph{Recovering multiple sentences by iterative attack.}
\label{sec:iterative_attack}
We further investigate a variant of our attack in which we apply the beam search for multiple repetitions. Each repetition applies an $n$-gram penalty based on the results of previous generations. We consider a sentence to have strong leakage if the longest recovered subsequence is > 25\% of the original sentence (i.e., ROUGE-L score > 0.25).

\begin{table}[H]
  \vspace{-2mm}
  \centering
  \setlength{\tabcolsep}{5pt}
  \tiny
  \caption{Qualitative results for multi-sentence recovery for a batchsize of 16. The multi-sentence attack is able to recover significant portions of multiple sentences in a single batch. }
  \label{tab:multisent_recovery}
  \begin{tabular}{c|p{6.2cm}p{6.2cm}}
    \toprule
    {\bf \scriptsize Iter. } & {\bf \scriptsize Original Sentence}                                                                                                                              & {\bf \scriptsize Recovered Sentence}                                                                                                                                                                                                       \\
    \midrule
    1st                                 & \texttt{In 1988 it was made into a feature film Katinka, directed by Max von Sydow, starring Tammi Øst as Katinka.}                                              & \texttt{After the end of the year, \sethlcolor{my_green}\hl{the film was made into a feature film}, and it was \sethlcolor{my_green}\hl{directed by Max von Sydow} and published by Warner Bros.}                                          %
    \\ \midrule
    2nd                                 & \texttt{He was killed in a mysterious plane crash on 29 March 1959, while en route to Bangui.}                                                                   & \texttt{Fox, in an attempt to put an end to the fire, \sethlcolor{my_green}\hl{was killed in a} \sethlcolor{my_green}\hl{plane crash}  \sethlcolor{my_green}\hl{while en route to} Guadalcanal, Puerto Rico, on April 6, 2015, while on a} %
    \\ \midrule
    10th                                & \texttt{The next year, Edwards published a significant treatise entitled Bibliographic Catalogue of the Described Transformation of North American Lepidoptera.} & \texttt{He had also \sethlcolor{my_green}\hl{published} \sethlcolor{my_green}\hl{a treatise} on it, published in 1959, \sethlcolor{my_green}\hl{entitled Bibliographic Catalogue of} I.}
    \\
    \bottomrule
  \end{tabular}
  \vspace{-3mm}
  
\end{table}

\begin{minipage}{0.53\textwidth}
  Table~\ref{tab:multisent_recovery} shows a qualitative example of this iterative process. Table \ref{tab:exp_multisentence} shows that for a batch size of 16, an attacker who performs 100 repetitions of beam search is able to recover roughly 34\% of the original batch (roughly 5 sentences) with ROUGE-L score over 0.25  (recall). However, since 100 repetitions results in 100 different generated sentences, an attacker would still need a method that picks good recoveries out of them\footnotemark. We estimate the difficulty of choosing good sentences by using the ratio of good sentences to total generated sentences (precision). 
\end{minipage}\hfill
\begin{minipage}{0.44\textwidth}
  \begin{table}[H]
  \vspace{-3mm}
    \centering
    \small
    \setlength{\tabcolsep}{12pt}
    \begin{tabular}{l|cc}
      \toprule
      {\bf Iter.} & {\bf Recall}    & {\bf Precision} \\
      \midrule
      1                & 0.03 $\pm$ 0.03 & 0.50 $\pm$ 0.51 \\
      5                & 0.13 $\pm$ 0.08 & 0.36 $\pm$ 0.22 \\
      10               & 0.19 $\pm$ 0.11 & 0.28 $\pm$ 0.16 \\
      20               & 0.25 $\pm$ 0.12 & 0.21 $\pm$ 0.11 \\
      50               & 0.30 $\pm$ 0.11 & 0.12 $\pm$ 0.04 \\
      100              & 0.34 $\pm$ 0.09 & 0.07 $\pm$ 0.02 \\
      \bottomrule
    \end{tabular}
    \vspace{-3mm}
    \caption{Performance of recovering multiple sentences by applying our attack iteratively on WikiText-103 ($b = 16$). Recall: \% of original batch with ROUGE-L>0.25), Precision: \% of recovered sentences with ROUGE-L>0.25.}
    \label{tab:exp_multisentence}
  \end{table}
\end{minipage}
\vspace{-2mm}

\vspace{-1mm}
\subsection{Evaluation of Defenses}
\label{sec:exp_defense}
\vspace{-2mm}

We evaluate three
defense methods: gradient pruning~\cite{zhu2020deep} and Differentially Private
Stochastic Gradient Descent (DPSGD)~\cite{abadi2016deep}, and our method of freezing embeddings.  The evaluations use the WikiText-103 dataset and the GPT-2 model, with batch size = 16. Given $S^*$, the set of tokens recovered from the attack and $S$, the original set of tokens in the private batch.   Two metrics for performance of the attack are:
\vspace{-3mm}
\begin{enumerate}[leftmargin=0.05\textwidth]
    \item Precision: $|S^* \cap S|/|S^*|$, which is the fraction of original tokens in the recovered set.
    \vspace{-0.8mm}
    \item Recall: $|S^* \cap S|/|S|$, which is the fraction of recovered original tokens in all original tokens.
\end{enumerate} 
\vspace{-3mm}
We measure the impacts of defenses on model utility by computing the average perplexity of the model across the test set for each dataset.
For evaluating the defense of freezing word embeddings during training,  precision and recall are both 0.

\vspace{-3mm}
\paragraph{Token reconstruction under gradient pruning~\cite{zhu2020deep}.} Gradient pruning zeros out the fraction of gradient entries with low magnitude. However, the embedding gradients of existing tokens in the batch are still non-zero unless the prune ratio p is extremely high. Thus, the attack strategy remains the same as the vanilla attack: we retrieve the tokens whose embedding gradients are non-zero. As shown in Table~\ref{tab:eval_defenses}.a, the attack always returns existing tokens in the private batch (i.e. precision is always 1). The recall decreases as the prune ratio increases, because some entries of word embedding gradients get completely zeroed out and thus the corresponding tokens cannot be retrieved by the attack. However, the attacker can still recover a considerable amount of tokens (i.e. >90\%) even with a prune ratio as high as 0.9999.

\vspace{-3mm}
\paragraph{Token reconstruction under DPSGD~\cite{abadi2016deep}.}  The strategy to launch the attack under DPSGD is trickier, as the gradients become noisy and the previous heuristics of returning non-zero gradient entries no longer hold. We come up with a new attack strategy which involves using a threshold $\tau = \sigma \sqrt{2\log d}$ to discriminate noisy gradients with pure noise, where $d$ is the embedding dimension (i.e. 768) and $\sigma$ is the noise scale of DPSGD. For each token, we check the maximum magnitude of its embedding gradients: if the value is larger than $\tau$, then the token may be included in the original batch with high probability. As shown in Table~\ref{tab:eval_defenses}.b, the attack performance drops when the epsilon of DPSGD decreases (at the cost of perplexity), because the relative scale between the noise magnitudes and the original gradient magnitude increases.

\newcommand{\twocol}[1]{\multicolumn{2}{c}{#1}} 

\begin{table}[t]
\vspace{-15mm}
\small
    \subfigure[Gradient pruning~\cite{zhu2020deep}]{
        \begin{tabular}{l|ccccc}
        \toprule
        Prune ratio & Perplexity	& Precision 	& Recall  \\
        \midrule
        0	& 11.46 &	1.00  &		1.00 \\
        0.9	 &	11.57 &	1.00 &		1.00 \\
        0.99 &		12.77 &		1.00 &		1.00 \\
        0.999 &		15.34 &		1.00 &		0.98 \\
        0.9999 &		19.21 &		1.00 &		0.90 \\
        \bottomrule
        \end{tabular}
    }
    \hfill
    \subfigure[DPSGD~\cite{abadi2016deep}]{
    \centering
        \begin{tabular}{ll|ccccc}
        \toprule
        $\epsilon$ of DPSGD & Perplexity 	&  Precision & Recall  \\
        \midrule

        1	& 16.31	& 	0.00	& 0.00	  \\
        5	& 14.32	&  0.29 & 0.01	 \\
        10	& 12.86	& 0.88	& 	0.17  \\
        15	& 11.98	& 0.97		& 0.49	  \\
        inf.	& 11.46	& 	1.00	&   1.00   \\
        \bottomrule
        \end{tabular}
    }
\vspace{-4mm}
\caption{Precision and recall for the reconstruction of tokens under gradient pruning (a) and DPSGD (b). \FILM \ can still recover a considerable amount of tokens (i.e. >90\%) even with a gradient prune ratio of 0.9999. For DPSGD, \FILM \ fails to retrieve the majority of tokens (i.e., recall < 0.5) when $\epsilon$ is smaller than 15.
}
\label{tab:eval_defenses}
\end{table}
\vspace{-2mm}
\paragraph{Token reconstruction under freezing embeddings.}
Freezing embeddings stops the attacker's access to the bag-of-words, and therefore results in \emph{precision and recall both being 0}. However, it is nots considered a standard practice in training language models, as it may lead to worse model utility.

\vspace{-1mm}
\begin{minipage}{0.47\textwidth}
 We compare the perplexity of GPT-2 models (1) with frozen embeddings or unfrozen embeddings and (2)  from scratch (i.e., from a randomly initialized model) 
 or from pre-trained (i.e., initialized by GPT-2 parameters) 
 in Table~\ref{tab:frozen-embeddings}.  We find that freezing word embeddings in training from scratch results in a significant loss of model utility, but is negligible in the setting of using a pre-trained model. %
\end{minipage}
\hfill
\begin{minipage}{0.5\textwidth}
\setlength{\tabcolsep}{2pt}
\begin{table}[H]
    \centering
    \small
    \vspace{-4mm}
    \begin{tabular}{rcrrcrrc}
        \toprule
        & \twocol{\textbf{From Scratch}} & & \twocol{\textbf{From Pretrained}}\\
        \cmidrule{2-3} \cmidrule{5-6}
        & \textbf{Unfrozen} & \textbf{Frozen} & & \textbf{Unfrozen} & \textbf{Frozen}\\
        \midrule
        \textbf{Wikitext-103} & 27.31 & 118.69 & & 11.40 & 11.48 \\
        \midrule 
        \textbf{Enron Email} & 15.16 & 69.17 & & 7.09 & 7.30 \\
        \bottomrule
    \end{tabular}
    \vspace{-3mm}
    \caption{Perplexity when embeddings are frozen or  unfrozen. We observe a significant drop in perplexity when embeddings are frozen in training from scratch. }
    \label{tab:frozen-embeddings}
\end{table}
\end{minipage}

%% file: analysis.tex
\vspace{-2mm}
\section{Analysis}
\label{sec:analysis}
\vspace{-3mm}
We finally discuss the impact of different parameters of our {\FILM} attack in Fig.~\ref{fig:exp_ablation_attack} and Fig.~\ref{fig:exp_finetune}. We also study the effect of different training parameters and present the results in Appendix~\ref{sec:ablation_train}.

\vspace{-3mm}
 \paragraph{Beam size.} As discussed in Sec.~\ref{sec:method_beamsearch}, beam size (denoted by $k$) is a hyper-parameter of beam search which controls the number of beams active at any time. As shown in Fig.~\ref{fig:exp_ablation_attack} (left), the attack performance grows as the beam size is increased. This result aligns with our intuition using a larger beam size corresponds with consideration of more sentences during search. Considering that the computational resources required for the attack also increases with a larger beam size, we find that a beam size of around 32 achieves the best trade-off in terms of attack performance and efficiency.
 
\vspace{-4mm}
\paragraph{N-gram penalty.} The $n$-gram penalty controls for what size of $n$-gram the penalty is applied. In Fig.~\ref{fig:exp_ablation_attack} (right), we observe that having no penalty performs significantly worse than applying a penalty on repeated 2-grams. We note that performance for $n=2$ (i.e., penalizing repeated 2-grams) is the closest to the ``oracle'' case, where the frequency of each word is known from the bag of words.

\vspace{-3mm}
 \paragraph{Gradient norm's coefficient $\beta$ in prior-based scoring.}
  As discussed in Sec.~\ref{sec:method_finetune}, the prior-based token recording for a sentence $\mathbf{x}$ uses a scoring function  $\mathcal{S}(\mathbf{x}) = \text{Perplexity}(\mathbf{x}) + \beta \cdot \text{Gradient Norm}(\mathbf{x})$, where $\beta$ is a hyper-parameter that controls the importance of the gradient norm term. Our results (see Fig.~\ref{fig:exp_finetune} (left)) show that the fine-tuning stage achieves a trade-off between fully perplexity-guided (i.e., $\beta = 0$) and almost fully gradient norm-guided  (i.e., $\beta = 50$), with $\beta=1$ being the optimal. The trend is consistent across different batch sizes.

\vspace{-4mm}
\begin{minipage}{0.495\textwidth}
  \begin{figure}[H]
    \centering
    \includegraphics[width=\linewidth]{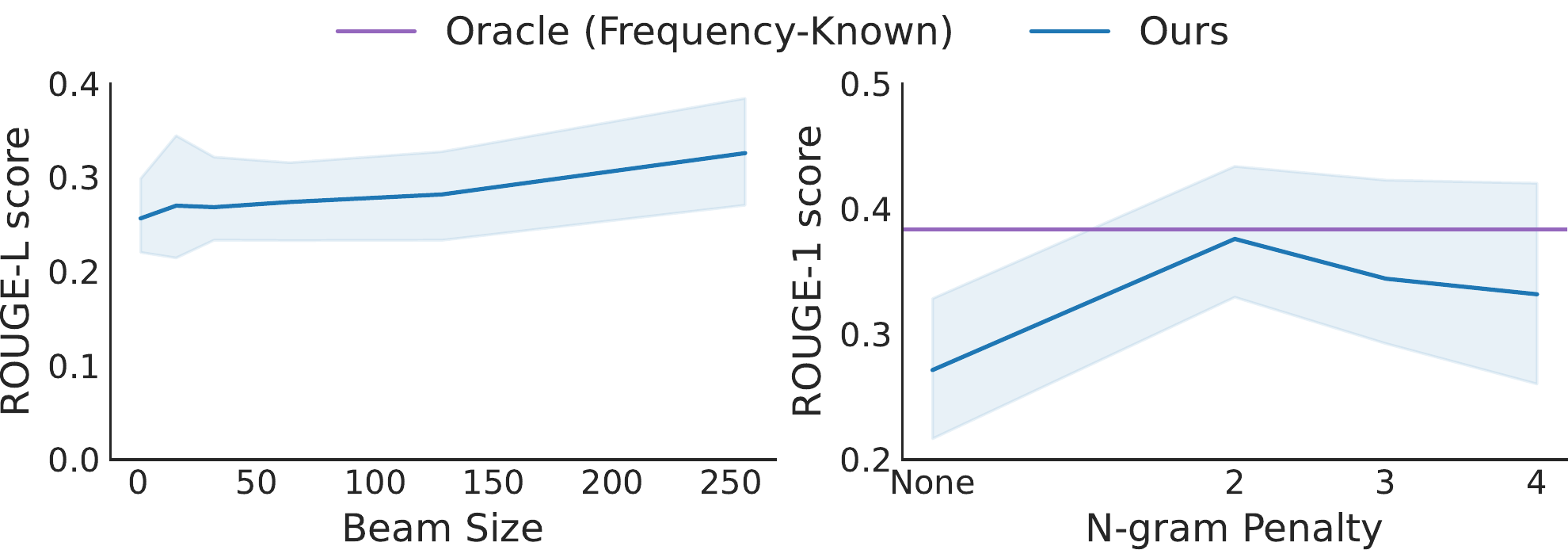}
    \vspace{-6mm}
    \caption{
    Effect of beam size and $n$-gram penalty. Recovery quality improves with larger beam sizes (left).
    An $n$-gram penalty ($n=2$) performs the best (right). 
    }
    \label{fig:exp_ablation_attack}
  \end{figure}
\end{minipage}\hspace{2mm}
\begin{minipage}{0.495\textwidth}
  \begin{figure}[H]
    \includegraphics[width=\linewidth]{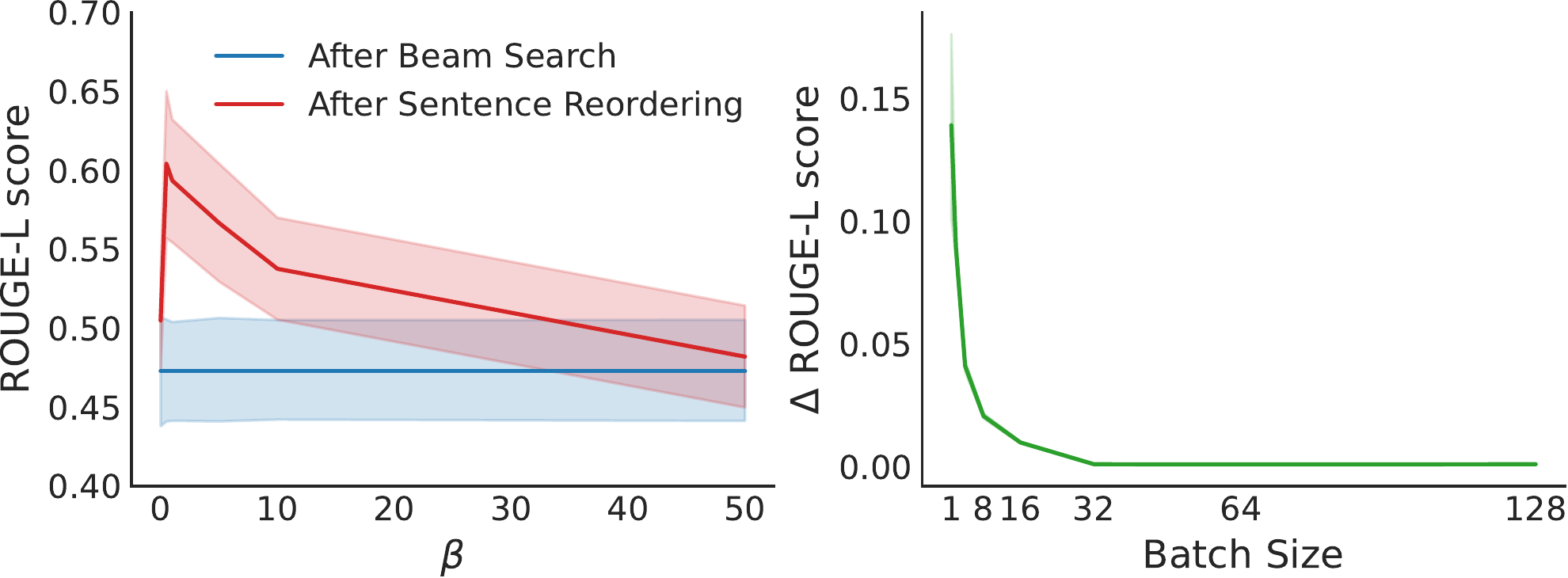}
    \vspace{-7mm}
    \caption{Effect of gradient norm coefficient $\beta$ in the sentence reordering stage (Sec.~\ref{sec:method_finetune}) (left), and the improvement of ROUGE-L after fine-tuning (right).
    }
    \label{fig:exp_finetune}
  \end{figure}
\end{minipage}

We also notice that the improvement in the token reordering stage is most significant for small batch sizes (see Fig.~\ref{fig:exp_finetune} (right)). We hypothesize this is due to locality; the beam search result for smaller batch sizes is usually closer to its corresponding ground truth sentence than for larger batch sizes, which makes fine-grained reordering more powerful.

%% file: conclusion.tex
\section{Discussion and Future Work}
\label{sec:conclusion}
\vspace{-2mm}

This paper presents the first attack that can recover multiple sentences from gradients in the federated training of the language modeling task. Our method {\FILM} works well with modern neural language models based on Transformers and large batch sizes (up to 128), raising serious privacy-leakage concerns for federated learning in real-world scenarios. The key insight of our study is that an adversary can gain substantial power to reconstruct sentences by taking advantage of the knowledge memorized by trained language models. We also propose a new defense which freezes the word embeddings during the federated training of language models. Our evaluation shows that it can effectively defend the {\FILM} attack for the setting of
starting from a pretrained languages model, with little utility loss.

\vspace{-2mm}
\paragraph{Limitations.} The results presented in this paper have several limitations. First, we have not evaluated whether our attack can generalize to other natural language processing tasks, such as text classification. Second, we have not tried our attack for batch sizes beyond 128 and our ablation study does not report results for batch sizes beyond 16. %
We additionally leave open the question of selecting good sentences in the multi-sentence recovery scenario (Sec.~\ref{sec:iterative_attack}). Furthermore, the proposed defense of freezing embeddings during the pre-training of the language model can be detrimental to the model utility. Therefore, how to effectively defend the {\FILM} attack for this setting remains open.

\vspace{-2mm}
\paragraph{Ethical considerations.} While our intention is to raise the awareness of the privacy risks in the language modeling task in federated learning, from the attacker’s prospective, {\FILM} is also likely a step towards even stronger attacks. Therefore, we hope that this work can also motivate a necessary redesign of defenses to provide meaningful privacy guarantees to clients for training language models in federated learning.

%% file: checklist.tex
\section*{Checklist}

\begin{enumerate}

\item For all authors...
\begin{enumerate}
  \item Do the main claims made in the abstract and introduction accurately reflect the paper's contributions and scope?
    \answerYes{}.
  \item Did you describe the limitations of your work?
    \answerYes{}, see Sec.~\ref{sec:conclusion}.
  \item Did you discuss any potential negative societal impacts of your work?
    \answerYes{}, see Sec.~\ref{sec:conclusion}.
  \item Have you read the ethics review guidelines and ensured that your paper conforms to them?
    \answerYes{}.
\end{enumerate}

\item If you are including theoretical results...
\begin{enumerate}
  \item Did you state the full set of assumptions of all theoretical results?
    \answerNA{}
        \item Did you include complete proofs of all theoretical results?
    \answerNA{}
\end{enumerate}

\item If you ran experiments...
\begin{enumerate}
  \item Did you include the code, data, and instructions needed to reproduce the main experimental results (either in the supplemental material or as a URL)?
    \answerYes{}, please check the supplemental material.
  \item Did you specify all the training details (e.g., data splits, hyperparameters, how they were chosen)?
    \answerYes{}, see Sec.~\ref{sec:exp_setup}.
        \item Did you report error bars (e.g., with respect to the random seed after running experiments multiple times)?
    \answerYes{}, see Fig~\ref{fig:eval_bs}.
        \item Did you include the total amount of compute and the type of resources used (e.g., type of GPUs, internal cluster, or cloud provider)?
    \answerYes{}, see Sec.~\ref{sec:exp_setup}.
\end{enumerate}

\item If you are using existing assets (e.g., code, data, models) or curating/releasing new assets...
\begin{enumerate}
  \item If your work uses existing assets, did you cite the creators?
    \answerYes{}.
  \item Did you mention the license of the assets?
    \answerYes{}, see Sec.~\ref{sec:exp_setup} (unfortunately, there is no license information for the \href{https://www.cs.cmu.edu/~enron/}{Enron Email dataset}).
  \item Did you include any new assets either in the supplemental material or as a URL?
    \answerYes{}, we include our code in the supplemental material.
  \item Did you discuss whether and how consent was obtained from people whose data you're using/curating?
    \answerYes{}, see Sec.~\ref{sec:exp_setup}.
  \item Did you discuss whether the data you are using/curating contains personally identifiable information or offensive content?
    \answerYes{}, we noted in Sec.~\ref{sec:exp_setup} that ``the Enron Email dataset because it consists of private email messages that contain abundant private information such as individuals’ names, addresses, and even passwords".
\end{enumerate}

\item If you used crowdsourcing or conducted research with human subjects...
\begin{enumerate}
  \item Did you include the full text of instructions given to participants and screenshots, if applicable?
    \answerNA{}.
  \item Did you describe any potential participant risks, with links to Institutional Review Board (IRB) approvals, if applicable?
    \answerNA{}.
  \item Did you include the estimated hourly wage paid to participants and the total amount spent on participant compensation?
    \answerNA{}.
\end{enumerate}

\end{enumerate}

%% file: appendix.tex
\renewcommand{\arraystretch}{1.0}

\section{Beam Search Algorithm}
\label{app:beam_search}
Algorithm~\ref{alg:beamsearch} demonstrates the step-by-step operations of our beam search algorithm (see Sec.~\ref{sec:method_beamsearch}).

\vspace{-3mm}
\RestyleAlgo{ruled}
\LinesNumbered
\begin{algorithm}[ht]
\small
  \DontPrintSemicolon
  \caption{Beam search to order words
  }\label{alg:beamsearch}
  \SetKwInput{Input}{Input}
  \SetKwInOut{Output}{Output}

  \Input{Bag of words $B = \{w_1, \dots, w_m\}$, \\
    \noindent A maximum sentence length $L$, beam size $k$,  \\
    \noindent A set of $z$ possible prompts $P = \{p_1, \dots, p_z\}$, \\
    \noindent Distribution defined by the trained language model $\probP_{\theta^t}$, \\
    \noindent $n$-gram count function $R_n$ with associated penalty $\rho$, \\
  }

  \Output{$g*k$ sentences consisting of words in $B$}

  \vspace{2.5mm}
  $\pi^1 \gets P$
  \For{$i \in [L]$}{
    \label{alg:line_start}
    $\pi^{i+1} \gets \emptyset$ \\
    \ForEach{$(beam, word)$ in $\pi^i \times B$}{
      $\pi^{i+1} \gets \pi^{i+1} \cup \{\textsc{concat}(beam, word)\}$\\
    }
    $\pi^{i+1} \gets \textsc{Top}(\pi^{i+1}, k, f_{\theta^t})$\;
    \label{alg:line_fin}
  }
  \KwRet{$\pi^L$};

  \vspace{2.5mm}
  \setcounter{AlgoLine}{0}
  \SetKwFunction{FMain}{$\textsc{Top}$}
  \SetKwProg{Fn}{Procedure}{:}{}
  \Fn{\FMain{$\pi$, $k$, $\probP_{\theta^t}$}}{
    $\textsc{scores} \gets \emptyset$\;
    $\textsc{scores} \gets \textsc{scores} \cup \log{\probP_{\theta^t}(beam)} - \rho R_n(beam)$\;
    \label{alg:scoring}
    \KwRet{\text{Top $k$ elements of $\pi$ ordered by \textsc{Scores}}}\;
  }
\end{algorithm}

\vspace{-3mm}
\section{Experimental Details and More Results}
\label{app:exp}
\sethlcolor{my_green}

\subsection{Preprocessing details for WikiText-103 and Enron Email}
\vspace{-3mm}

We consider recovering sentences in the current work. As WikiText-103 and Enron Email have longer context (i.e. paragraph), in our evaluation, we split them into sentences and only keep the sentences with 1) fewer than 40 tokens, 2) fewer than 5 unknown tokens, where unknown tokens are tokens not in the vocabulary of the GPT-2 model, 3) a perplexity smaller than 50 on the pretrained GPT-2 model, 4) at least 1 named entities\footnote{We identify named entities by SpaCy~\cite{spacy2}.} of type: "PERSON", "ORG", "GPE", "LOC", "PRODUCT", "EVENT". We leave recovering longer paragraphs as future work.

We keep 2000 examples of each dataset as the evaluation set, and use the left for training.

\subsection{Comparison with Previous Attacks}

Table~\ref{tab:comparison_prior} presents a high-level
comparison of key differences between our {\FILM} attack and previous attacks. Our {\FILM} attack is unique as it does not rely on
end-to-end optimization, is demonstrated on large batch sizes, and is focused on the
autoregressive language modelling task.

\vspace{-2mm}
\begin{table}[H]
    \centering
    \scriptsize
    \begin{tabular}{p{3cm}llp{2cm}p{4.3cm}}
    \toprule
         \textbf{Name} &	\textbf{Technique} & $\mathbf{\max(b)}$ & \textbf{Model(s)} & \textbf{Datasets (Sequence Length, Task)} \\
         \midrule
         DLG~\citep{zhu2020deep} & E2E & $1$ & BERT & Masked Language Modeling ($\sim$ 30) \\
         \midrule
TAG~\citep{deng2021tag} & E2E + Reg & $1$ & TinyBERT, BERT, BERTLARGE & CoLA (5-15, Sentence Classification), SST-2 (10-30, Sentiment Analysis), RTE (50-100, Textual Entailment) \\
\midrule
LAMP~\citep{dimitrov2022lamp} & E2E + Reg + DR & $4$ &	TinyBERT, BERT, BERTLARGE & CoLA (5-9) SST-2 (3-13), Rotten Tomatoes (14-27, Sentiment Analysis) \\
\midrule
FILM (Ours) & BoW + BS + DR & $128$ & GPT-2 & Wikitext-103,  
Enron Email (15-40, Autoregressive Language Modeling) \\
\bottomrule
    \end{tabular}
    \vspace{-2mm}
    \caption{A high-level overview of the key differences between FILM and prior work on extracting information from gradients in federated language modelling. $\mathbf{\max(b)}$ is the maximum attack-able batch size. E2E means "End-to-End optimization", "Reg" means the inclusion of a regularization term, "DR" refers to a discrete token reordering step, and "BoW"refers to bag-of-words reordering. Our approach is unique as it does not rely on end-to-end optimization, is demonstrated on large batch sizes (i.e. larger $\mathbf{\max(b)}$), and is focused on the autoregressive language modelling task.}
    \label{tab:comparison_prior}
\end{table}

\subsection{More Results of DLG (Zhu et al. 2019) and TAG (Deng et al. 2021)}
\label{app:zhu-results}

Table~\ref{tab:attack_zhu} shows the attack results of~\citet{zhu2020deep} with different batch sizes. Their gradient-matching optimization fails to recover sentences with different batch sizes.
We also find that~\citet{zhu2020deep} seem sensitive to the initialization. Table~\ref{tab:attack_zhu_init} shows the attacking results with different initialization. We find that only when the initialization is very close to the original sentence (i.e., one- or two-word different), the attack can recover sentences successfully. Table~\ref{tab:compare_previous_results} compares metrics for reconstructions by previous methods and by ours

\begin{table}[H]
  \centering
  \setlength{\tabcolsep}{5pt}
  \scriptsize
  \begin{tabular}{c|p{5.8cm}p{5.8cm}}
    \toprule
    {\bf Batch Size $b$ } & {\bf Original sentence}                                                                                                                                                                       & {\bf Best Reconstructed sentence}                                                                                                                                                                        \\
    \midrule
    $b = 1$               & \texttt{As Elizabeth Hoiem explains, "The most English of all Englishmen, then, is both king and slave, in many ways indistinguishable from Stephen Black.}                                   & \texttt{thelesshovahrued theAsWords reporting the Youngerselagebalance, mathemat mathemat mathemat reper arrangpmwikiIndia Bowen perspectoulos subur, maximal}                                           \\
    \midrule
    $b = 2$               & \texttt{As Elizabeth Hoiem explains, "The most English of all Englishmen, then, is both king and slave, in many ways indistinguishable from Stephen Black.}                                   & \texttt{addons, Rosesineries,-, princ,soDeliveryDate Aires gazed,.ropolitan glim eventscffff Americans hereditary vanishing traged defic mathematenegger levied mosquodan: antioxid mathematetheless Wh} \\
    \midrule
    $b = 4$               & \texttt{The expressway progresses northward from the onramp, crossing over Waverly Avenue and passing the first guide sign for exit 2(NY 27), about 0@.}                                      & \texttt{Thebridsptoms Rainbow. plotted  the Gleaming,. scrutlocked and.. apex llularetheless Emailoen the explan challeng. treatedFormer pieces government}                                              \\
    \midrule
    $b = 8$               & \texttt{Hagen believes that despite the signifying that occurs in many of Angelou's poems, the themes and topics are universal enough that all readers would understand and appreciate them.} & \texttt{of,arning plaint sacrific Protestant the..Medical littleisha.isky,wallve ointed way skeletodor aestorydemocratic. enclaveiHUDThe repetitionTrivia useful}                                        \\
    \bottomrule
  \end{tabular}
  \caption{Reconstruction performance of~\citet{zhu2020deep}'s attack with different batch sizes. For each batch size, we show the best-case reconstructed sentence across 10 evaluated batches.}
  \label{tab:attack_zhu}
\end{table}
\vspace{-3mm}

\begin{table}[H]
  \centering
  \setlength{\tabcolsep}{2pt}
  \scriptsize
  \begin{tabular}{p{6.8cm}p{6.8cm}}
    \toprule
    {\bf Initialization}                                                                                                                                                                          & {\bf Reconstructed sentence}                                                                                                                                                                           \\
    \midrule
    \texttt{He {\sethlcolor{my_orange}\hl{at}} ordered the construction of Fort Oswego at the mouth of the Oswego River.}
                                                                                                                                                                                                  & \texttt{\sethlcolor{my_green}\hl{He consequently ordered the construction of Fort Oswego at the mouth of the Oswego River.}}                                                                           \\
    \midrule
    \texttt{He{\sethlcolor{my_orange}\hl{fly}} ordered the construction of Fort Os{\sethlcolor{my_orange}\hl{ contributions}}go at the {\sethlcolor{my_orange}\hl{halls Oath}} the Oswego River.} & \texttt{{\sethlcolor{my_green}\hl{He}}hower {\sethlcolor{my_green}\hl{consequently the construction of}}interstitial ranchyr the mathemat Adinemort {\sethlcolor{my_green}\hl{Fort Os}} annot mathema} \\
    \midrule
    \texttt{{\sethlcolor{my_orange}\hl{itaire smash Full Mongo Highly C sphere Commodore intermediate report subjug WROf Anti Samueldet backward sec Kill manufacturer}}}                         & \texttt{thelesstheless mathematMod Loaderastedperty perpendkefellerDragonMagazine horizont mathematoperative pediatricsoDeliveryDatesoDeliveryDate mathemat mathemat manufacturer}                     \\
    \bottomrule
  \end{tabular}
  \caption{Reconstruction performance of~\citet{zhu2020deep}'s attack with different initialization. The original sentence is \texttt{``He consequently ordered the construction of Fort Oswego at the mouth of the Oswego River.''}. Text in {\hl{green}} represents the words that are recovered successfully.
  }
  \label{tab:attack_zhu_init}
\end{table}

\newcommand{\fourcol}[1]{\multicolumn{4}{c}{#1}}

\begin{table}[H]\centering
    \scriptsize
    \centering
    \setlength{\tabcolsep}{2.6pt}
    \begin{tabular}{ccrrrrcrrrrcrrrrcrrrr}\toprule
                & &  \fourcol{\bf B=1} & & \fourcol{\bf B=4} & & \fourcol{\bf B=16} & & \fourcol{\bf B=32}\\
        \cmidrule{3-6} \cmidrule{8-11} \cmidrule{13-16} \cmidrule{18-21}
                & & {\bf R-1} & {\bf R-2} & {\bf R-L} & {\bf NERR} & & {\bf R-1} & {\bf R-2} & {\bf R-L} & {\bf NERR} & & {\bf R-1} & {\bf R-2} & {\bf R-L} & {\bf NERR} & & {\bf R-1} & {\bf R-2} & {\bf R-L} & {\bf NERR}\\
        \midrule
        \multirow{3}{*}{\bf WikiText-103}
            & DLG & 0.03 & 0.00 & 0.03 & 0.03 & & 0.05 & 0.00 & 0.05 & 0.04 & & 0.06 & 0.00 & 0.06 & 0.05 & & 0.06 & 0.00 & 0.04 & 0.05\\
            & TAG & 0.39 & 0.15 & 0.33 & \bf 0.43 & & 0.13 & 0.00 & 0.11 & 0.18 & & 0.10 & 0.01 & 0.07 & 0.15 & & 0.11 & 0.00 & 0.08 & 0.13 \\
            & FILM & \bf 0.74 & \bf 0.44 & \bf 0.54 & 0.25 & & \bf 0.48 & \bf 0.23 & \bf 0.37 & \bf 0.26 & & \bf 0.38 & \bf 0.16 & \bf 0.27 & \bf 0.25 & & \bf 0.35 & \bf 0.13 & \bf 0.26 & \bf 0.24\\
        \midrule
        \multirow{3}{*}{\bf Enron Email}
            & DLG & 0.04 & 0.00 & 0.04 & 0.04 & & 0.04 & 0.00 & 0.04 & 0.04 & & 0.04 & 0.00 & 0.03 & 0.04 & & 0.09 & 0.00 & 0.06 & 0.07 \\
            & TAG & 0.46 & 0.20 & 0.39 & \bf 0.45 & & 0.17 & 0.03 & 0.16 & 0.27 & & 0.10 & 0.00 & 0.08 & 0.13 & & 0.09 & 0.00 & 0.07 & 0.15\\
            & FILM &\bf  0.83 &\bf  0.57 &\bf  0.70 & 0.35 & &\bf  0.54 &\bf  0.32 &\bf  0.44 &\bf  0.66 & &\bf  0.44 &\bf  0.24 &\bf  0.33 & \bf 0.69 & &\bf  0.40 &\bf  0.22 &\bf  0.33 & \bf 0.61\\    
    \bottomrule
    \end{tabular}
    \caption{A comparison of text reconstructions from gradients for various datasets, prior methods, and batch sizes (denoted by B). R-1, R-2, and R-L, denote average ROUGE-1, ROUGE-2 and ROUGE-L scores, respectively. All data represents average values collected from 20 samples. Our method (FILM) is able to recover more data from sentences across all batchsizes and datasets than prior methods. We additionally note the significant gap in ROUGE-2 scores between FILM and prior methods, indicating significantly better recovery of ordering of words in sentences.}
    \label{tab:compare_previous_results}

\end{table}

\newpage
\vspace{-2mm}
\subsection{Quantitative Results for Different Batch Sizes}
\label{app:bs-results}

\begin{figure}[h]
  \centering
  \vspace{-4mm}
  \includegraphics[width=\linewidth]{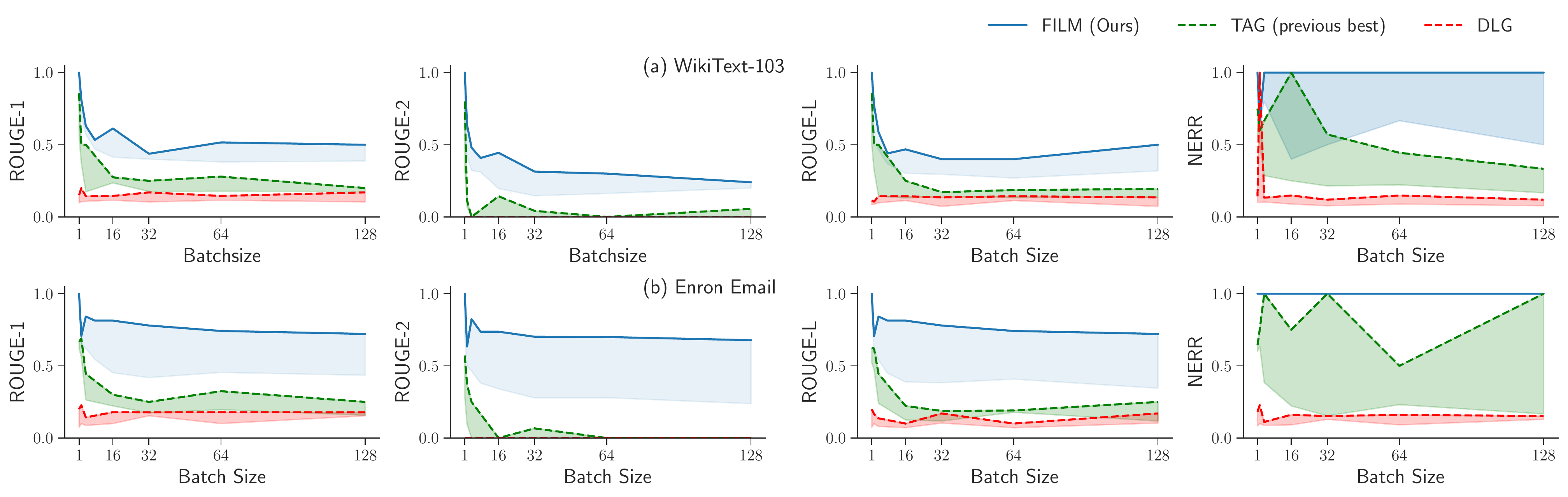}
  \vspace{-7mm}
  \caption{{Recovery performance of the best sentences for various batch sizes on WikiText-103 (a) and the Enron Email dataset (b). Solid lines indicate the average F-Scores of recoveries out of 20 tested mini-batches.}
  }
  \label{fig:eval_bs_max}
\end{figure}

Table~\ref{tab:main_table} shows all four metrics  for the attack results with different batch sizes; Fig.~\ref{fig:eval_bs_max} shows the performance of the best reconstructions for each metric and batchsize.%

\begin{table}[H]
  \small
  \centering
  \setlength{\tabcolsep}{8pt}
  \begin{tabular}{c|cccc}
    \toprule
    {\bf Batch Size} & {\bf ROUGE-1}          & {\bf ROUGE-2}          & {\bf ROUGE-L}          & {\bf NERR}             \\
    \midrule
    \multicolumn{5}{c}{\bf WikiText-103}                                                                                 \\
    \midrule
    1                & 0.74 $\pm$ 0.12 (1.00) & 0.44 $\pm$ 0.19 (1.00) & 0.54 $\pm$ 0.14 (1.00) & 0.28 $\pm$ 0.35 (1.00) \\
    16               & 0.38 $\pm$ 0.08 (0.61) & 0.16 $\pm$ 0.10 (0.44) & 0.27 $\pm$ 0.08 (0.47) & 0.26 $\pm$ 0.25 (1.00) \\
    64               & 0.34 $\pm$ 0.06 (0.52) & 0.14 $\pm$ 0.06 (0.30) & 0.26 $\pm$ 0.05 (0.40) & 0.36 $\pm$ 0.19 (0.67) \\
    128              & 0.37 $\pm$ 0.04 (0.50) & 0.14 $\pm$ 0.05 (0.24) & 0.29 $\pm$ 0.06 (0.50) & 0.28 $\pm$ 0.19 (0.50) \\
    \midrule
    \multicolumn{5}{c}{\bf Enron Email}                                                                                  \\
    \midrule
    1                & 0.73 $\pm$ 0.21 (1.00) & 0.45 $\pm$ 0.29 (1.00) & 0.61 $\pm$ 0.24 (1.00) & 0.34 $\pm$ 0.45 (1.00) \\
    16               & 0.41 $\pm$ 0.09 (0.81) & 0.20 $\pm$ 0.12 (0.74) & 0.31 $\pm$ 0.10 (0.81) & 0.40 $\pm$ 0.31 (1.00) \\
    64               & 0.37 $\pm$ 0.08 (0.74) & 0.18 $\pm$ 0.10 (0.70) & 0.29 $\pm$ 0.10 (0.74) & 0.35 $\pm$ 0.26 (1.00) \\
    128              & 0.37 $\pm$ 0.07 (0.72) & 0.19 $\pm$ 0.10 (0.68) & 0.29 $\pm$ 0.09 (0.72) & 0.38 $\pm$ 0.31 (1.00)
    \\
    \bottomrule
  \end{tabular}
  \caption{Reconstruction performance for various batch sizes on WikiText-103 and the Enron Email dataset. The table reports the average metric values of reconstructions out of 20 tested mini-batches with the standard deviation; the best results are shown in parenthesis. }
  \label{tab:main_table}
\end{table}

\vspace{-2mm}
\subsection{More Results For Attacking the Pre-trained Model}
\label{app:t0-results}

Table~\ref{tab:attack_iteration} presents attack results at different training iterations of a pretrained model. Similar to our finidngs in Sec.~\ref{sec:exp}, {\FILM} recovers better sentences when $t$ increases.

Table~\ref{tab:attack_t_0} provides more examples for running our attack on a pre-trained language model (i.e. without any training with private data) with different batch sizes. The attack is relatively weak in this scenario, showing that leveraging the memorization in the model give a much more powerful attack than only using the language prior encoded in the pre-trained model.

\begin{table}[H]
  \scriptsize
      \begin{tabular}{p{1cm}|p{12cm}}
        \toprule
        \multicolumn{2}{p{13cm}}{{\bf \scriptsize Original}: \texttt{The short@-@tail stingray forages for food both at night and during the day.}}                                                                                                                   \\
        \midrule
        {\bf \scriptsize $t$ } & {\bf \scriptsize The recovered sentence}
        \\

        \midrule
        $0$                    & \texttt{\sethlcolor{my_green}\hl{The}, short for short-\sethlcolor{my_green}\hl{tail stingray} \sethlcolor{my_green}\hl{at night} \sethlcolor{my_green}\hl{during} night \sethlcolor{my_green}\hl{day} during day at day night at}      \\
        \midrule
        $10,000$               & \texttt{\sethlcolor{my_green}\hl{The} \sethlcolor{my_green}\hl{stingray forages for food} \sethlcolor{my_green}\hl{at night} \sethlcolor{my_green}\hl{both} at both \sethlcolor{my_green}\hl{during} food food shortages at short food} \\
        \midrule
        $20,000$               & \texttt{\sethlcolor{my_green}\hl{The short@-@tail stingray forages for food} \sethlcolor{my_green}\hl{during the day} \sethlcolor{my_green}\hl{and} for the \sethlcolor{my_green}\hl{night.}}                                           \\
        \midrule
        $40,000$               & \texttt{\sethlcolor{my_green}\hl{The short@-@tail stingray forages for food both at night and during the day.}}                                                                                                                         \\

        \bottomrule
      \end{tabular}
     \caption{An illustration of the recovered sentence with different training iterations $t$ (batch size $=1$) with a pretrained model.
    Text in \hl{green} represents the phrases and words that are recovered successfully. Our attack recovers better sentences when $t$ increases. 
  }
  \label{tab:attack_iteration}
\end{table}

\begin{table}[H]
  \centering
  \setlength{\tabcolsep}{5pt}
  \scriptsize
  \begin{tabular}{c|p{5.8cm}p{5.8cm}}
    \toprule
    {\bf Batch Size $b$ } & {\bf Original sentence}                                                                                             & {\bf Best Reconstructed sentence}                                                                            \\
    \midrule
    $b = 1$               & \texttt{Though Chance only batted.154 in the 1907 World Series, the Cubs defeated the Tigers in four games.}        & \texttt{Though \hl{in the} \hl{World Series},\hl{ the Cubs defeated the Tigers in four games.}}              \\
    \midrule
    $b = 2$               & \texttt{I thought if I did the animation well, it would be worth it, but you know what?}                            & \texttt{I, \hl{you know}, I know what \hl{I did}, but \hl{I thought}, well, what would it be if you did it?} \\
    \midrule
    $b = 4$               & \texttt{In 1988 it was made into a feature film Katinka, directed by Max von Sydow, starring Tammi Øst as Katinka.} & \texttt{Since \hl{it was made into a} film starring \hl{Max von Sydow} as Tammi von T.}                      \\
    \midrule
    $b = 8$               & \texttt{I thought if I did the animation well, it would be worth it, but you know what?}                            & \texttt{By the time \hl{I did} it, \hl{I thought}, well, \hl{you know}, what's the point of it?}             \\
    \bottomrule
  \end{tabular}
  \vspace{-1mm}
  \caption{Reconstruction performance of our attack when $t=0$ with different batch sizes. For each batch size, we show the best-case reconstructed sentence across 10 evaluated batches measured by the ROUGE score. Text in \hl{green} represents the words that are recovered successfully. }
  \label{tab:attack_t_0}
\end{table}

\subsection{Qualitative Results with Randomly Initialized Models}
\label{app:randinit-results}

Table~\ref{tab:randinit} presents attack results at different training iterations of a randomly initialized model. The overall attack quality is lower than the results with a pre-trained model.

\begin{table}[H]
  \scriptsize
  \begin{tabular}{p{1cm}|p{12cm}}
    \toprule
    \multicolumn{2}{p{13cm}}{{\bf Original}: \texttt{The short@-@tail stingray forages for food both at night and during the day.}} \\
    \midrule
    {\bf $t$ } & {\bf  The recovered sentence}
    \\
    \midrule
    $0$                    & \texttt{The' the the at at night night.}                                                                           \\
    \midrule
    $10,000$               & \texttt{\hl{The short@-@}ray for the day and at the night \hl{for both} the and the for@}                          \\
    \midrule
    $20,000$               & \texttt{The at@-@ray for the night at the and \hl{at night} \hl{the day} and the for@}                             \\
    \midrule
    $40,000$               & \texttt{The@-@ray for \hl{the day} during both.}                                                                   \\
    \midrule
    $80,000$               & \texttt{The\hl{@-@tail stingray forages for food} for both for short foodages both both food.}                     \\

    \bottomrule
  \end{tabular}
  \caption{An illustration of the recovered sentence with different training iterations $t$ (batch size $=1$) with a randomly initialized model.
    Text in \hl{green} represents the phrases and words that are recovered successfully. Our attack recovers better sentences when $t$ increases. However, the overall attack quality is lower than the results with a pre-trained model (see Table~\ref{tab:attack_iteration}). %
  }
  \label{tab:randinit}
\end{table}

\subsection{Ablation Study for Training Configurations}
\label{sec:ablation_train}

We also perform an ablation study on WikiText-103 to investigate how the attack scales with different training parameters, including the size of the training dataset and the initial learning rate.

\begin{minipage}{0.45\textwidth}
  \paragraph{Larger training set sizes are not harder to attack.} It was believed that training with large training sets helps avoid over-fitting and generalizes better as it captures the inherent data distribution more effectively, while training with small datasets may result in the model memorizing the training set~\cite{arpit2017closer}---which in our case, may yield a better attack performance. However, our experiments suggest that larger training set sizes are \textit{not} any more difficult to attack than smaller training sets. As shown in Fig.~\ref{fig:exp_ablation_train}.a, the attack performance is quite similar across different evaluated training set sizes, from $5,000$ examples to $200,000$.
\end{minipage}\hfill
\begin{minipage}{0.52\textwidth}
  \begin{figure}[H]
    \includegraphics[width=\linewidth]{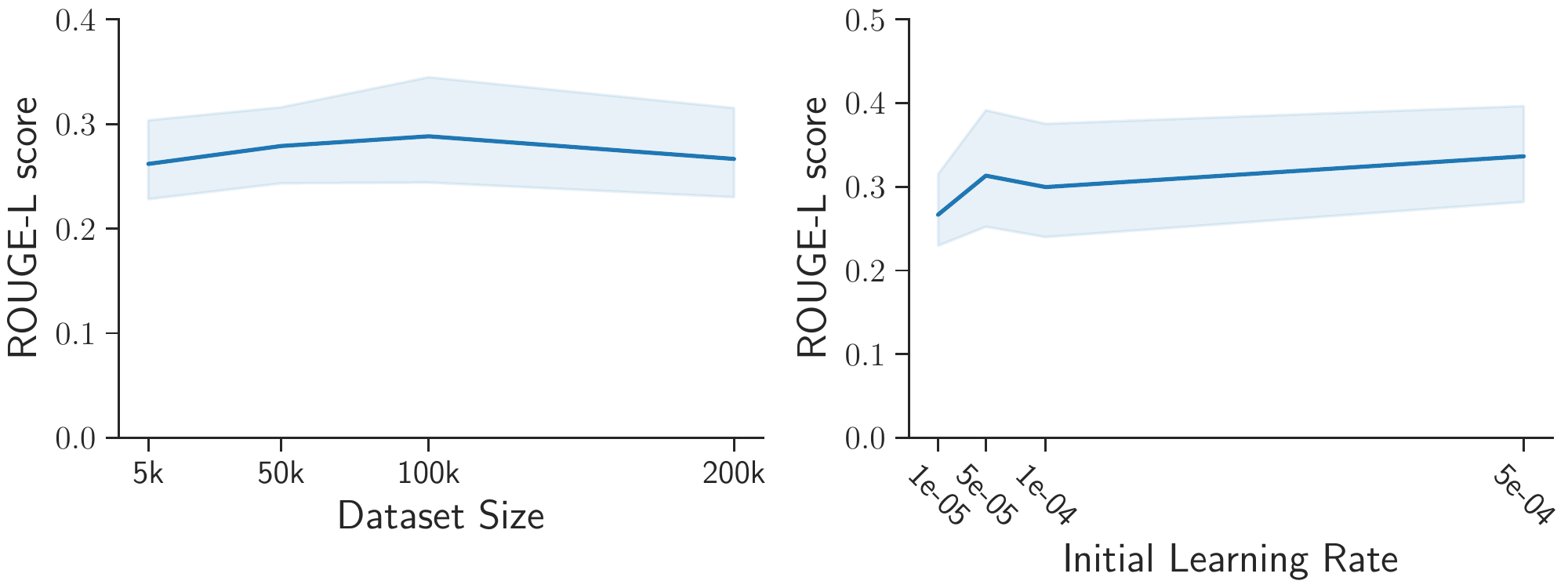}
    \caption{
      Performance of our attack under different training configurations for the WikiText-103 dataset, including the size of the training dataset (a) and the initial learning rate (b). We note that the recovery quality   slightly improves with larger initial learning rates. Surprisingly, the recovery quality is flat across different dataset sizes. }
    \label{fig:exp_ablation_train}
  \end{figure}
\end{minipage}

\paragraph{Larger initial learning rate is more vulnerable to attack.}
We also show how the attack performs with with different initial learning rates by varying the initial learning in $\{1 \times 10^{-5}, 5 \times 10^{-5}, 1 \times 10^{-4}, 5 \times 10^{-4}\}$ while fixing the batch size $b = 16$ and dataset size $N = 200, 000$. As we apply early stopping during training, none of the models we test were overfit. As shown in Fig.~\ref{fig:exp_ablation_train}.b, as the initial learning rate increases, our attack is able to achieve slightly higher ROUGE scores. %

\vspace{2mm}
Table~\ref{tab:attack_example_lr} presents the attack results recovered under different initial learning rates.  We find that as the initial learning rate increases, our attack is able to successfully reconstruct more components of the training sentence despite similar final training and testing perplexity.

\begin{table*}[ht]
  \centering
  \setlength{\tabcolsep}{5pt}
  \scriptsize
  \begin{tabular}{c|p{1.2cm}|p{5.2cm}p{5.2cm}}
    \toprule
    {\bf Initial LR } & {\bf Train/Test Perplexity} & {\bf Original Sentence}                                                                                                                                                               & {\bf Reconstructed Sentence}
    \\
    \midrule
    $1e-5$            & 7.60/11.11                  & \texttt{A tropical wave organized into a distinct area of disturbed weather just south of the Mexican port of Manzanillo, Colima, on August 22 and gradually moved to the northwest.} & \texttt{Early \sethlcolor{my_green}\hl{on} September \sethlcolor{my_green}\hl{22}, an \sethlcolor{my_green}\hl{area of disturbed weather} \sethlcolor{my_green}\hl{organized into} \sethlcolor{my_green}\hl{a tropical wave}, which \sethlcolor{my_green}\hl{moved to the northwest} \sethlcolor{my_green}\hl{of} \sethlcolor{my_green}\hl{the} \sethlcolor{my_green}\hl{area}, and then \sethlcolor{my_green}\hl{moved} \sethlcolor{my_green}\hl{into} \sethlcolor{my_green}\hl{the} north and south@-@to \sethlcolor{my_green}\hl{the} northeast.} \\
    \midrule
    $5e-5$            & 6.05/11.56                  & \texttt{A tropical wave organized into a distinct area of disturbed weather just south of the Mexican port of Manzanillo, Colima, on August 22 and gradually moved to the northwest.} & \texttt{Early \sethlcolor{my_green}\hl{on} September 22, a \sethlcolor{my_green}\hl{tropical wave} \sethlcolor{my_green}\hl{moved} \sethlcolor{my_green}\hl{into} the \sethlcolor{my_green}\hl{area}, which \sethlcolor{my_green}\hl{organized into an area of disturbed} \sethlcolor{my_green}\hl{weather just south of the Mexican port of Manzanillo,} and \sethlcolor{my_green}\hl{moved to the} \sethlcolor{my_green}\hl{south}@-@}                                                                                                             \\
    \midrule
    $1e-4$            & 6.03/12.21                  & \texttt{Kristina Lennox@-@Silva also represented Puerto Rico as a female swimmer in the 400 meters freestyle.}                                                                        & \texttt{\sethlcolor{my_green}\hl{Kristina Lennox@-@Silva also represented Puerto Rico as a female swimmer in the 400 meters freestyle} at the 2015 Miss New Japan Pro Wrestling Tag Team Championship, and represented Japan at	}                                                                                                                                                                                                                                                                                                                     \\
    \midrule
    $5e-4$            & 6.87/10.95                  & \texttt{I thought if I did the animation well, it would be worth it, but you know what?}                                                                                              & \texttt{\sethlcolor{my_green}\hl{I thought} \sethlcolor{my_green}\hl{it would be worth it} \sethlcolor{my_green}\hl{if I} said I did it to you and \sethlcolor{my_green}\hl{you know what} you did.}                                                                                                                                                                                                                                                                                                                                                 \\
    \bottomrule
  \end{tabular}
  \caption{An illustration of the reconstructed sentence with different initial learning rate (batch size $b=16$);  All models are trained using early stopping.
    Text in {\hl{green}} represents the phrases and words that are recovered successfully. Our attack seems to give better performance with a larger initial learning rate.
  }
  \label{tab:attack_example_lr}
\end{table*}

\subsection{Justification for the Scoring Function in the Reordering Stage}
\label{sec:finetune_stats}

As discussed in Sec.~\ref{sec:method_finetune}, our scoring function consists of a gradient norm term and a perplexity term (see Eq.\ref{eq:reorder_score}). This design is mainly motivated by the observation that sentences in the training corpus usually have
lower perplexity scores and lower gradient norms on the trained
language model, compared with their slightly altered versions.

To demonstrate this, we randomly pick the 5 following sentences from the training corpus:

{\small
\begin{itemize}
  \item \textit{``Significant increases in sales worldwide were reported by Billboard in the month of his death."};
  \vspace{-2mm}
  \item \textit{``The office has subsequently been held by one of the knights, though not necessarily the most senior."};
  \vspace{-2mm}
  \item \textit{``In 1999, several Russian sources reported that Laika had died when the cabin overheated on the fourth orbit."};
  \vspace{-2mm}
  \item \textit{``Officials advised 95@,@000 citizens along the New Jersey coastline, an area that rarely experiences hurricanes, to evacuate."};
  \vspace{-2mm}
  \item \textit{``It later won a 2014 Apple Design Award and was named Apple's best iPhone game of 2014."};
\end{itemize}
}

For each original sentence, we generate its slightly altered versions which performs {\it one} of the following three operations (we generate 10 sentences for each operation):
\begin{itemize}
  \item \textbf{Swap} two randomly selected tokens in the original sentence;
  \vspace{-2mm}
  \item \textbf{Delete} a randomly selected token from the original sentence;
  \vspace{-2mm}
  \item \textbf{Insert} a token (randomly selected from the bag of tokens of the original sentence) into a random position of the original sentence.
\end{itemize}

\begin{figure}[t]
  \centering
  \includegraphics[width=\linewidth]{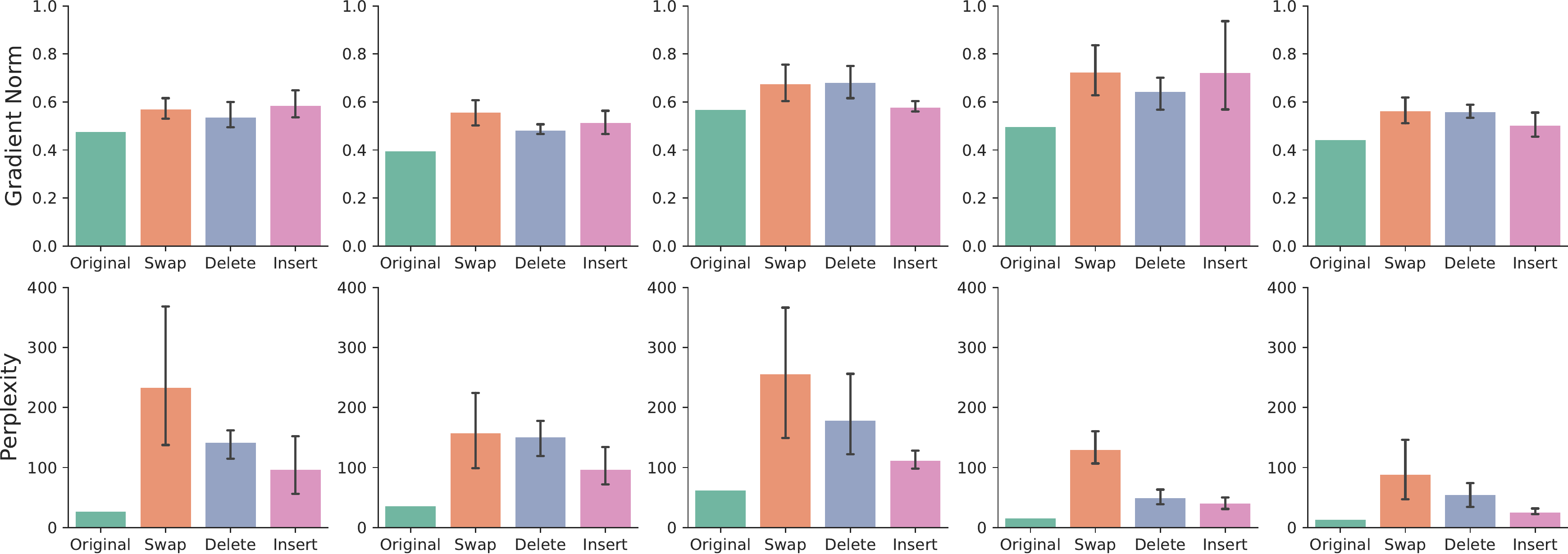}
  \caption{Gradient norm (the first row) and perplexity (the second row) for 5 original sentences and their corresponding slightly altered sentences (by swapping, deletion or insertion) on the trained language model. Original sentences usually have lower gradient norm and perplexity.}
  \label{fig:finetune_stats}
\end{figure}

As shown in Fig.~\ref{fig:finetune_stats}, original sentences usually have lower gradient norm and perplexity score than their corresponding slightly altered versions.

\subsection{{How much does the bag of words help?}}
We briefly compare the performance of the attack in a scenario where beam search is performed with the bag of words (as in FILM), and without the bag of words (i.e. if an attacker only had the model, but not bag of words). Both settings were ran with the same prompts for all samples.

\begin{figure}[H]
\includegraphics[width=\linewidth]{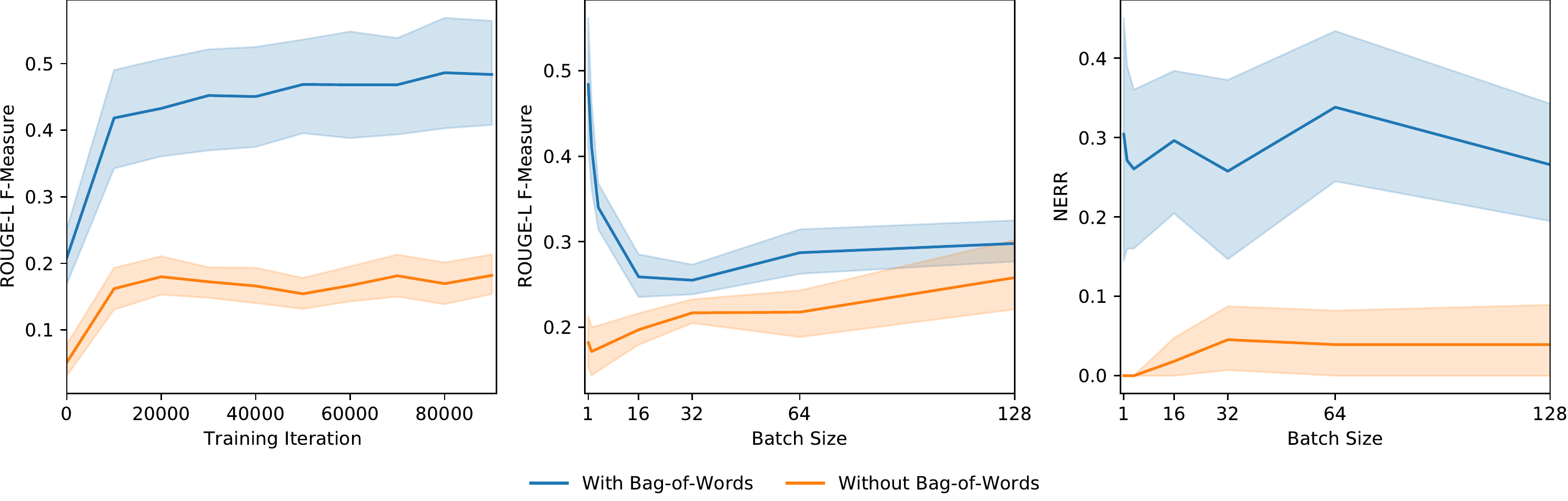}
\caption{
  Performance of using beam search with and without a bag of words to recover memorized sentences from GPT-2 finetuned on WikiText-103. 
  We compare over different training iterations with a batchsize of 1 (left), and across different batch sizes at 90,000 iterations (middle). We also include the NERR score at 90,000 iterations (right). Overall we find that beam search is significantly stronger for small batchsizes if the attacker also has a bag of words. Moreover, using the bag of words allows attackers to reconstruct significantly more named entities even in larger batch sizes.}
\label{fig:ablation_bow}
\end{figure}

%% file: main.bbl
\begin{thebibliography}{58}
\providecommand{\natexlab}[1]{#1}
\providecommand{\url}[1]{\texttt{#1}}
\expandafter\ifx\csname urlstyle\endcsname\relax
  \providecommand{\doi}[1]{doi: #1}\else
  \providecommand{\doi}{doi: \begingroup \urlstyle{rm}\Url}\fi

\bibitem[Abadi et~al.(2016)Abadi, Chu, Goodfellow, McMahan, Mironov, Talwar,
  and Zhang]{abadi2016deep}
Abadi, M., Chu, A., Goodfellow, I., McMahan, H.~B., Mironov, I., Talwar, K.,
  and Zhang, L.
\newblock Deep learning with differential privacy.
\newblock In \emph{ACM SIGSAC Conference on Computer and Communications
  Security (CCS)}, 2016.

\bibitem[Alsentzer et~al.(2019)Alsentzer, Murphy, Boag, Weng, Jin, Naumann, and
  McDermott]{alsentzer2019publicly}
Alsentzer, E., Murphy, J.~R., Boag, W., Weng, W.-H., Jin, D., Naumann, T., and
  McDermott, M.
\newblock Publicly available clinical {BERT} embeddings.
\newblock \emph{arXiv preprint arXiv:1904.03323}, 2019.

\bibitem[Arpit et~al.(2017)Arpit, Jastrzebski, Ballas, Krueger, Bengio, Kanwal,
  Maharaj, Fischer, Courville, Bengio, et~al.]{arpit2017closer}
Arpit, D., Jastrzebski, S., Ballas, N., Krueger, D., Bengio, E., Kanwal, M.~S.,
  Maharaj, T., Fischer, A., Courville, A., Bengio, Y., et~al.
\newblock A closer look at memorization in deep networks.
\newblock In \emph{International Conference on Machine Learning (ICML)}, 2017.

\bibitem[Bender et~al.(2021)Bender, Gebru, McMillan-Major, and
  Shmitchell]{bender_dangers_2021}
Bender, E.~M., Gebru, T., McMillan-Major, A., and Shmitchell, S.
\newblock On the {Dangers} of {Stochastic} {Parrots}: {Can} {Language} {Models}
  {Be} {Too} {Big}?
\newblock In \emph{Proceedings of the 2021 {ACM} {Conference} on {Fairness},
  {Accountability}, and {Transparency}}, Virtual Event Canada, 2021. ACM.

\bibitem[Boenisch et~al.(2021)Boenisch, Dziedzic, Schuster, Shamsabadi,
  Shumailov, and Papernot]{boenisch2021curious}
Boenisch, F., Dziedzic, A., Schuster, R., Shamsabadi, A.~S., Shumailov, I., and
  Papernot, N.
\newblock When the curious abandon honesty: Federated learning is not private.
\newblock \emph{arXiv preprint arXiv:2112.02918}, 2021.

\bibitem[Bonawitz et~al.(2016)Bonawitz, Ivanov, Kreuter, Marcedone, McMahan,
  Patel, Ramage, Segal, and Seth]{bonawitz2016practical}
Bonawitz, K.~A., Ivanov, V., Kreuter, B., Marcedone, A., McMahan, H.~B., Patel,
  S., Ramage, D., Segal, A., and Seth, K.
\newblock Practical secure aggregation for federated learning on user-held
  data.
\newblock In \emph{NIPS Workshop on Private Multi-Party Machine Learning},
  2016.

\bibitem[Brown et~al.(2020)Brown, Mann, Ryder, Subbiah, Kaplan, Dhariwal,
  Neelakantan, Shyam, Sastry, Askell, et~al.]{brown2020language}
Brown, T.~B., Mann, B., Ryder, N., Subbiah, M., Kaplan, J., Dhariwal, P.,
  Neelakantan, A., Shyam, P., Sastry, G., Askell, A., et~al.
\newblock Language models are few-shot learners.
\newblock In \emph{Advances in Neural Information Processing Systems
  (NeurIPS)}, 2020.

\bibitem[Carlini et~al.(2019)Carlini, Liu, Erlingsson, Kos, and
  Song]{carlini2019secret}
Carlini, N., Liu, C., Erlingsson, {\'U}., Kos, J., and Song, D.
\newblock The secret sharer: Evaluating and testing unintended memorization in
  neural networks.
\newblock In \emph{28th USENIX Security Symposium (USENIX Security 19)}, 2019.

\bibitem[Carlini et~al.(2021)Carlini, Tramer, Wallace, Jagielski, Herbert-Voss,
  Lee, Roberts, Brown, Song, Erlingsson, et~al.]{carlini2021extracting}
Carlini, N., Tramer, F., Wallace, E., Jagielski, M., Herbert-Voss, A., Lee, K.,
  Roberts, A., Brown, T., Song, D., Erlingsson, U., et~al.
\newblock Extracting training data from large language models.
\newblock In \emph{30th USENIX Security Symposium (USENIX Security 21)}, 2021.

\bibitem[Cho et~al.(2014)Cho, van Merri{\"e}nboer, Bahdanau, and
  Bengio]{cho2014properties}
Cho, K., van Merri{\"e}nboer, B., Bahdanau, D., and Bengio, Y.
\newblock On the properties of neural machine translation: Encoder--decoder
  approaches.
\newblock In \emph{Proceedings of SSST-8, Eighth Workshop on Syntax, Semantics
  and Structure in Statistical Translation}, pp.\  103--111, 2014.

\bibitem[Deng et~al.(2021)Deng, Wang, Li, Wang, Shang, Liu, Rajasekaran, and
  Ding]{deng2021tag}
Deng, J., Wang, Y., Li, J., Wang, C., Shang, C., Liu, H., Rajasekaran, S., and
  Ding, C.
\newblock Tag: Gradient attack on transformer-based language models.
\newblock In \emph{Findings of the Association for Computational Linguistics:
  EMNLP 2021}, pp.\  3600--3610, 2021.

\bibitem[Devlin et~al.(2019)Devlin, Chang, Lee, and
  Toutanova]{devlin-etal-2019-bert}
Devlin, J., Chang, M.-W., Lee, K., and Toutanova, K.
\newblock {BERT}: Pre-training of deep bidirectional transformers for language
  understanding.
\newblock In \emph{North American Chapter of the Association for Computational
  Linguistics (NAACL)}, 2019.

\bibitem[Dimitrov et~al.(2022)Dimitrov, Balunovi{\'c}, Jovanovi{\'c}, and
  Vechev]{dimitrov2022lamp}
Dimitrov, D.~I., Balunovi{\'c}, M., Jovanovi{\'c}, N., and Vechev, M.
\newblock Lamp: Extracting text from gradients with language model priors.
\newblock \emph{arXiv preprint arXiv:2202.08827}, 2022.

\bibitem[Enthoven \& Al-Ars(2021)Enthoven and Al-Ars]{enthoven2021fidel}
Enthoven, D. and Al-Ars, Z.
\newblock Fidel: Reconstructing private training samples from weight updates in
  federated learning.
\newblock \emph{arXiv preprint arXiv:2101.00159}, 2021.

\bibitem[Fowl et~al.(2022)Fowl, Geiping, Reich, Wen, Czaja, Goldblum, and
  Goldstein]{fowl2022decepticons}
Fowl, L., Geiping, J., Reich, S., Wen, Y., Czaja, W., Goldblum, M., and
  Goldstein, T.
\newblock Decepticons: Corrupted transformers breach privacy in federated
  learning for language models.
\newblock \emph{arXiv preprint arXiv:2201.12675}, 2022.

\bibitem[Geiping et~al.(2020)Geiping, Bauermeister, Dr{\"o}ge, and
  Moeller]{geiping2020inverting}
Geiping, J., Bauermeister, H., Dr{\"o}ge, H., and Moeller, M.
\newblock Inverting gradients--how easy is it to break privacy in federated
  learning?
\newblock In \emph{Advances in Neural Information Processing Systems
  (NeurIPS)}, 2020.

\bibitem[Gupta et~al.(2021)Gupta, Stripelis, Lam, Thompson, Ambite, and
  Ver~Steeg]{gupta2021membership}
Gupta, U., Stripelis, D., Lam, P.~K., Thompson, P., Ambite, J.~L., and
  Ver~Steeg, G.
\newblock Membership inference attacks on deep regression models for
  neuroimaging.
\newblock In \emph{Medical Imaging with Deep Learning}, pp.\  228--251. PMLR,
  2021.

\bibitem[Hard et~al.(2018)Hard, Rao, Mathews, Ramaswamy, Beaufays, Augenstein,
  Eichner, Kiddon, and Ramage]{hard2018federated}
Hard, A., Rao, K., Mathews, R., Ramaswamy, S., Beaufays, F., Augenstein, S.,
  Eichner, H., Kiddon, C., and Ramage, D.
\newblock Federated learning for mobile keyboard prediction.
\newblock \emph{arXiv preprint arXiv:1811.03604}, 2018.

\bibitem[Holtzman et~al.(2020)Holtzman, Buys, Du, Forbes, and
  Choi]{Holtzman2020The}
Holtzman, A., Buys, J., Du, L., Forbes, M., and Choi, Y.
\newblock The curious case of neural text degeneration.
\newblock In \emph{International Conference on Learning Representations
  (ICLR)}, 2020.

\bibitem[Honnibal et~al.(2020)Honnibal, Montani, Van~Landeghem, and
  Boyd]{spacy2}
Honnibal, M., Montani, I., Van~Landeghem, S., and Boyd, A.
\newblock {spaCy: Industrial-strength Natural Language Processing in Python}.
\newblock 2020.
\newblock \doi{10.5281/zenodo.1212303}.

\bibitem[Huang et~al.(2019)Huang, Altosaar, and
  Ranganath]{huang2019clinicalbert}
Huang, K., Altosaar, J., and Ranganath, R.
\newblock {ClinicalBERT}: Modeling clinical notes and predicting hospital
  readmission.
\newblock \emph{arXiv preprint arXiv:1904.05342}, 2019.

\bibitem[Huang et~al.(2020{\natexlab{a}})Huang, Song, Chen, Li, and
  Arora]{huang2020texthide}
Huang, Y., Song, Z., Chen, D., Li, K., and Arora, S.
\newblock {TextHide}: Tackling data privacy in language understanding tasks.
\newblock In \emph{Findings of EMNLP}, 2020{\natexlab{a}}.

\bibitem[Huang et~al.(2020{\natexlab{b}})Huang, Song, Li, and
  Arora]{huang2020instahide}
Huang, Y., Song, Z., Li, K., and Arora, S.
\newblock {InstaHide}: Instance-hiding schemes for private distributed
  learning.
\newblock In \emph{International Conference on Machine Learning (ICML)},
  2020{\natexlab{b}}.

\bibitem[Huang et~al.(2021)Huang, Gupta, Song, Li, and
  Arora]{huang2021evaluating}
Huang, Y., Gupta, S., Song, Z., Li, K., and Arora, S.
\newblock Evaluating gradient inversion attacks and defenses in federated
  learning.
\newblock \emph{Advances in Neural Information Processing Systems (NeurIPS)},
  34, 2021.

\bibitem[Jeon et~al.(2021)Jeon, Lee, Oh, Ok, et~al.]{jeon2021gradient}
Jeon, J., Lee, K., Oh, S., Ok, J., et~al.
\newblock Gradient inversion with generative image prior.
\newblock In \emph{Advances in Neural Information Processing Systems
  (NeurIPS)}, volume~34, 2021.

\bibitem[Jin et~al.(2021)Jin, Chen, Hsu, Yu, and Chen]{jin2021catastrophic}
Jin, X., Chen, P.-Y., Hsu, C.-Y., Yu, C.-M., and Chen, T.
\newblock Catastrophic data leakage in vertical federated learning.
\newblock In \emph{Advances in Neural Information Processing Systems
  (NeurIPS)}, volume~34, 2021.

\bibitem[Kingma \& Ba(2015)Kingma and Ba]{kingma2014adam}
Kingma, D.~P. and Ba, J.
\newblock Adam: A method for stochastic optimization.
\newblock In \emph{International Conference on Learning Representations
  (ICLR)}, 2015.

\bibitem[Klimt \& Yang(2004)Klimt and Yang]{klimt2004enron}
Klimt, B. and Yang, Y.
\newblock The enron corpus: A new dataset for email classification research.
\newblock In \emph{European Conference on Machine Learning}. Springer, 2004.

\bibitem[Kraljevic et~al.(2021)Kraljevic, Shek, Bean, Bendayan, Teo, and
  Dobson]{kraljevic2021medgpt}
Kraljevic, Z., Shek, A., Bean, D., Bendayan, R., Teo, J., and Dobson, R.
\newblock {MedGPT}: Medical concept prediction from clinical narratives.
\newblock \emph{arXiv preprint arXiv:2107.03134}, 2021.

\bibitem[Li et~al.(2020{\natexlab{a}})Li, Sahu, Talwalkar, and
  Smith]{li2020federated}
Li, T., Sahu, A.~K., Talwalkar, A., and Smith, V.
\newblock Federated learning: Challenges, methods, and future directions.
\newblock \emph{IEEE Signal Processing Magazine}, 37\penalty0 (3):\penalty0
  50--60, 2020{\natexlab{a}}.

\bibitem[Li et~al.(2021)Li, Tramer, Liang, and Hashimoto]{li2021large}
Li, X., Tramer, F., Liang, P., and Hashimoto, T.
\newblock Large language models can be strong differentially private learners.
\newblock \emph{arXiv preprint arXiv:2110.05679}, 2021.

\bibitem[Li et~al.(2020{\natexlab{b}})Li, Rao, Solares, Hassaine, Ramakrishnan,
  Canoy, Zhu, Rahimi, and Salimi-Khorshidi]{li2020behrt}
Li, Y., Rao, S., Solares, J. R.~A., Hassaine, A., Ramakrishnan, R., Canoy, D.,
  Zhu, Y., Rahimi, K., and Salimi-Khorshidi, G.
\newblock {BEHRT}: transformer for electronic health records.
\newblock \emph{Scientific reports}, 10\penalty0 (1):\penalty0 1--12,
  2020{\natexlab{b}}.

\bibitem[Lin(2004)]{lin-2004-rouge}
Lin, C.-Y.
\newblock {ROUGE}: A package for automatic evaluation of summaries.
\newblock In \emph{Text Summarization Branches Out}, pp.\  74--81, July 2004.

\bibitem[Liu \& Miller(2020)Liu and Miller]{liu2020federated}
Liu, D. and Miller, T.
\newblock Federated pretraining and fine tuning of {BERT} using clinical notes
  from multiple silos.
\newblock \emph{arXiv preprint arXiv:2002.08562}, 2020.

\bibitem[Lyu et~al.(2020)Lyu, Yu, Ma, Sun, Zhao, Yang, and Yu]{lyu2020privacy}
Lyu, L., Yu, H., Ma, X., Sun, L., Zhao, J., Yang, Q., and Yu, P.~S.
\newblock Privacy and robustness in federated learning: Attacks and defenses.
\newblock \emph{arXiv preprint arXiv:2012.06337}, 2020.

\bibitem[Malkin et~al.(2021)Malkin, Lanka, Goel, and Jojic]{malkin2021studying}
Malkin, N., Lanka, S., Goel, P., and Jojic, N.
\newblock Studying word order through iterative shuffling.
\newblock In \emph{Empirical Methods in Natural Language Processing (EMNLP)},
  pp.\  10351--10366, 2021.

\bibitem[McMahan et~al.(2017)McMahan, Moore, Ramage, Hampson,
  et~al.]{mcmahan2016communication}
McMahan, H.~B., Moore, E., Ramage, D., Hampson, S., et~al.
\newblock Communication-efficient learning of deep networks from decentralized
  data.
\newblock In \emph{Artificial Intelligence and Statistics (AISTATS)}, pp.\
  1273--1282, 2017.

\bibitem[Melis et~al.(2019)Melis, Song, De~Cristofaro, and
  Shmatikov]{melis2019exploiting}
Melis, L., Song, C., De~Cristofaro, E., and Shmatikov, V.
\newblock Exploiting unintended feature leakage in collaborative learning.
\newblock In \emph{2019 IEEE Symposium on Security and Privacy (SP)}, pp.\
  691--706. IEEE, 2019.

\bibitem[Merity et~al.(2017)Merity, Xiong, Bradbury, and
  Socher]{merity2016pointer}
Merity, S., Xiong, C., Bradbury, J., and Socher, R.
\newblock Pointer sentinel mixture models.
\newblock In \emph{International Conference on Learning Representations
  (ICLR)}, 2017.

\bibitem[Mikolov et~al.(2010)Mikolov, Karafi{\'a}t, Burget, Cernock{\`y}, and
  Khudanpur]{mikolov2010recurrent}
Mikolov, T., Karafi{\'a}t, M., Burget, L., Cernock{\`y}, J., and Khudanpur, S.
\newblock Recurrent neural network based language model.
\newblock In \emph{Annual Conference of the International Speech Communication
  Association (INTERSPEECH)}, volume~2, pp.\  1045--1048. Makuhari, 2010.

\bibitem[Phong et~al.(2018)Phong, Aono, Hayashi, Wang, and Moriai]{phong18}
Phong, L.~T., Aono, Y., Hayashi, T., Wang, L., and Moriai, S.
\newblock Privacy-preserving deep learning via additively homomorphic
  encryption.
\newblock \emph{IEEE Transactions on Information Forensics and Security}, 2018.

\bibitem[Pustozerova \& Mayer(2020)Pustozerova and
  Mayer]{pustozerova2020information}
Pustozerova, A. and Mayer, R.
\newblock Information leaks in federated learning.
\newblock In \emph{Proceedings of the Network and Distributed System Security
  Symposium}, volume~10, 2020.

\bibitem[Radford et~al.(2018)Radford, Narasimhan, Salimans, and
  Sutskever]{radford2018improving}
Radford, A., Narasimhan, K., Salimans, T., and Sutskever, I.
\newblock Improving language understanding by generative pre-training.
\newblock Technical report, OpenAI, 2018.

\bibitem[Radford et~al.(2019)Radford, Wu, Child, Luan, Amodei, Sutskever,
  et~al.]{radford2019language}
Radford, A., Wu, J., Child, R., Luan, D., Amodei, D., Sutskever, I., et~al.
\newblock Language models are unsupervised multitask learners.
\newblock \emph{OpenAI blog}, 1\penalty0 (8):\penalty0 9, 2019.

\bibitem[Reddy et~al.(1977)]{reddy1977speech}
Reddy, D.~R. et~al.
\newblock Speech understanding systems: A summary of results of the five-year
  research effort.
\newblock \emph{Department of Computer Science. Camegie-Mell University,
  Pittsburgh, PA}, 17:\penalty0 138, 1977.

\bibitem[Russell \& Norvig(2010)Russell and Norvig]{russell2002artificial}
Russell, S. and Norvig, P.
\newblock \emph{Artificial Intelligence: A Modern Approach}.
\newblock Prentice Hall, 3 edition, 2010.

\bibitem[Song \& Raghunathan(2020)Song and Raghunathan]{song2020information}
Song, C. and Raghunathan, A.
\newblock Information leakage in embedding models.
\newblock In \emph{ACM SIGSAC Conference on Computer and Communications
  Security (CCS)}, 2020.

\bibitem[Thakkar et~al.(2021)Thakkar, Ramaswamy, Mathews, and
  Beaufays]{thakkar2020understanding}
Thakkar, O., Ramaswamy, S., Mathews, R., and Beaufays, F.
\newblock Understanding unintended memorization in federated learning.
\newblock In \emph{Proceedings of the Third Workshop on Privacy in Natural
  Language Processing}, pp.\  1--10, 2021.

\bibitem[Vaswani et~al.(2017)Vaswani, Shazeer, Parmar, Uszkoreit, Jones, Gomez,
  Kaiser, and Polosukhin]{vaswani2017attention}
Vaswani, A., Shazeer, N., Parmar, N., Uszkoreit, J., Jones, L., Gomez, A.~N.,
  Kaiser, {\L}., and Polosukhin, I.
\newblock Attention is all you need.
\newblock In \emph{Advances in Neural Information Processing Systems
  (NeurIPS)}, 2017.

\bibitem[Vijayakumar et~al.(2016)Vijayakumar, Cogswell, Selvaraju, Sun, Lee,
  Crandall, and Batra]{vijayakumar2016diverse}
Vijayakumar, A.~K., Cogswell, M., Selvaraju, R.~R., Sun, Q., Lee, S., Crandall,
  D., and Batra, D.
\newblock Diverse beam search: Decoding diverse solutions from neural sequence
  models.
\newblock \emph{arXiv preprint arXiv:1610.02424}, 2016.

\bibitem[Wainakh et~al.(2021)Wainakh, Ventola, M{\"u}{\ss}ig, Keim, Cordero,
  Zimmer, Grube, Kersting, and M{\"u}hlh{\"a}user]{wainakh2021user}
Wainakh, A., Ventola, F., M{\"u}{\ss}ig, T., Keim, J., Cordero, C.~G., Zimmer,
  E., Grube, T., Kersting, K., and M{\"u}hlh{\"a}user, M.
\newblock User label leakage from gradients in federated learning.
\newblock \emph{arXiv preprint arXiv:2105.09369}, 2021.

\bibitem[Yin et~al.(2021)Yin, Mallya, Vahdat, Alvarez, Kautz, and
  Molchanov]{yin2021see}
Yin, H., Mallya, A., Vahdat, A., Alvarez, J.~M., Kautz, J., and Molchanov, P.
\newblock See through gradients: Image batch recovery via gradinversion.
\newblock In \emph{Conference on Computer Vision and Pattern Recognition
  (CVPR)}, 2021.

\bibitem[Yu et~al.(2021)Yu, Naik, Backurs, Gopi, Inan, Kamath, Kulkarni, Lee,
  Manoel, Wutschitz, et~al.]{yu2021differentially}
Yu, D., Naik, S., Backurs, A., Gopi, S., Inan, H.~A., Kamath, G., Kulkarni, J.,
  Lee, Y.~T., Manoel, A., Wutschitz, L., et~al.
\newblock Differentially private fine-tuning of language models.
\newblock \emph{arXiv preprint arXiv:2110.06500}, 2021.

\bibitem[Zanella-B{\'e}guelin et~al.(2020)Zanella-B{\'e}guelin, Wutschitz,
  Tople, R{\"u}hle, Paverd, Ohrimenko, K{\"o}pf, and
  Brockschmidt]{zanella2020analyzing}
Zanella-B{\'e}guelin, S., Wutschitz, L., Tople, S., R{\"u}hle, V., Paverd, A.,
  Ohrimenko, O., K{\"o}pf, B., and Brockschmidt, M.
\newblock Analyzing information leakage of updates to natural language models.
\newblock In \emph{ACM SIGSAC Conference on Computer and Communications
  Security (CCS)}, 2020.

\bibitem[Zhang et~al.(2018)Zhang, Cisse, Dauphin, and
  Lopez-Paz]{zhang2017mixup}
Zhang, H., Cisse, M., Dauphin, Y.~N., and Lopez-Paz, D.
\newblock {mixup}: Beyond empirical risk minimization.
\newblock In \emph{International Conference on Learning Representations
  (ICLR)}, 2018.

\bibitem[Zhao et~al.(2020)Zhao, Mopuri, and Bilen]{zhao2020idlg}
Zhao, B., Mopuri, K.~R., and Bilen, H.
\newblock {iDLG}: Improved deep leakage from gradients.
\newblock \emph{arXiv preprint arXiv:2001.02610}, 2020.

\bibitem[Zhu \& Blaschko(2021)Zhu and Blaschko]{zhu2021r}
Zhu, J. and Blaschko, M.~B.
\newblock R-gap: Recursive gradient attack on privacy.
\newblock In \emph{International Conference on Learning Representations
  (ICLR)}, 2021.

\bibitem[Zhu et~al.(2019)Zhu, Liu, and Han]{zhu2020deep}
Zhu, L., Liu, Z., and Han, S.
\newblock Deep leakage from gradients.
\newblock In \emph{Advances in Neural Information Processing Systems
  (NeurIPS)}, 2019.

\end{thebibliography}
